\documentclass[sigconf]{acmart}

\AtBeginDocument{%
  \providecommand\BibTeX{{%
    \normalfont B\kern-0.5em{\scshape i\kern-0.25em b}\kern-0.8em\TeX}}}

\copyrightyear{2022} 
\acmYear{2022} 
\setcopyright{acmcopyright}
\acmConference[MM '22]{Proceedings of the 30th ACM International Conference on Multimedia}{October 10--14, 2022}{Lisboa, Portugal} 
\acmBooktitle{Proceedings of the 30th ACM International Conference on Multimedia (MM '22), October 10--14, 2022, Lisboa, Portugal} 
\acmPrice{15.00} 
\acmDOI{10.1145/3503161.3548410}
\acmISBN{978-1-4503-9203-7/22/10} 

\acmSubmissionID{3090}

\usepackage{multirow}

\begin{document}

\title{Source-Free Domain Adaptation for Real-World Image Dehazing}

\author{Hu Yu}
\authornotemark[1]
\affiliation{
   \institution{University of Science and Technology of China}
}
\email{yuhu520@mail.ustc.edu.cn}

\author{Jie Huang}
\authornote{Equal contribution.}
\affiliation{
   \institution{University of Science and Technology of China}
}
\email{hj0117@mail.ustc.edu.cn}

\author{Yajing Liu}
\affiliation{
   \institution{JD Logistics}
 }
\email{liuyajing25@jd.com}

\author{Qi Zhu}
\affiliation{
   \institution{University of Science and Technology of China}
 }
\email{zqcrafts@mail.ustc.edu.cn}

\author{Man Zhou}
\affiliation{%
   \institution{University of Science and Technology of China}
 }
\email{manman@mail.ustc.edu.cn}

\author{Feng Zhao}
\authornote{Corresponding author.}
\affiliation{%
   \institution{University of Science and Technology of China}
 }
\email{fzhao956@ustc.edu.cn}

\begin{abstract}
Deep learning-based source dehazing methods trained on synthetic datasets have achieved remarkable performance but suffer from dramatic performance degradation on real hazy images due to domain shift.
Although certain Domain Adaptation (DA) dehazing methods have been presented, they inevitably require access to the source dataset to reduce the gap between the source synthetic and target real domains.
To address these issues, we present a novel Source-Free Unsupervised Domain Adaptation (SFUDA) image dehazing paradigm, in which only a well-trained source model and an unlabeled target real hazy dataset are available.
Specifically, we devise the Domain Representation Normalization (DRN) module to make the representation of real hazy domain features match that of the synthetic domain to bridge the gaps. With our plug-and-play DRN module, unlabeled real hazy images can adapt existing well-trained source networks.
Besides, the unsupervised losses are applied to guide the learning of the DRN module, which consists of frequency losses and physical prior losses. Frequency losses provide structure and style constraints, while the prior loss explores the inherent statistic property of haze-free images.
Equipped with our DRN module and unsupervised loss, existing source dehazing models are able to dehaze unlabeled real hazy images.
Extensive experiments on multiple baselines demonstrate the validity and superiority of our method visually and quantitatively.

\end{abstract}




\maketitle

\section{Introduction}

\begin{figure}[t]
  \centering
  \includegraphics[width=\linewidth]{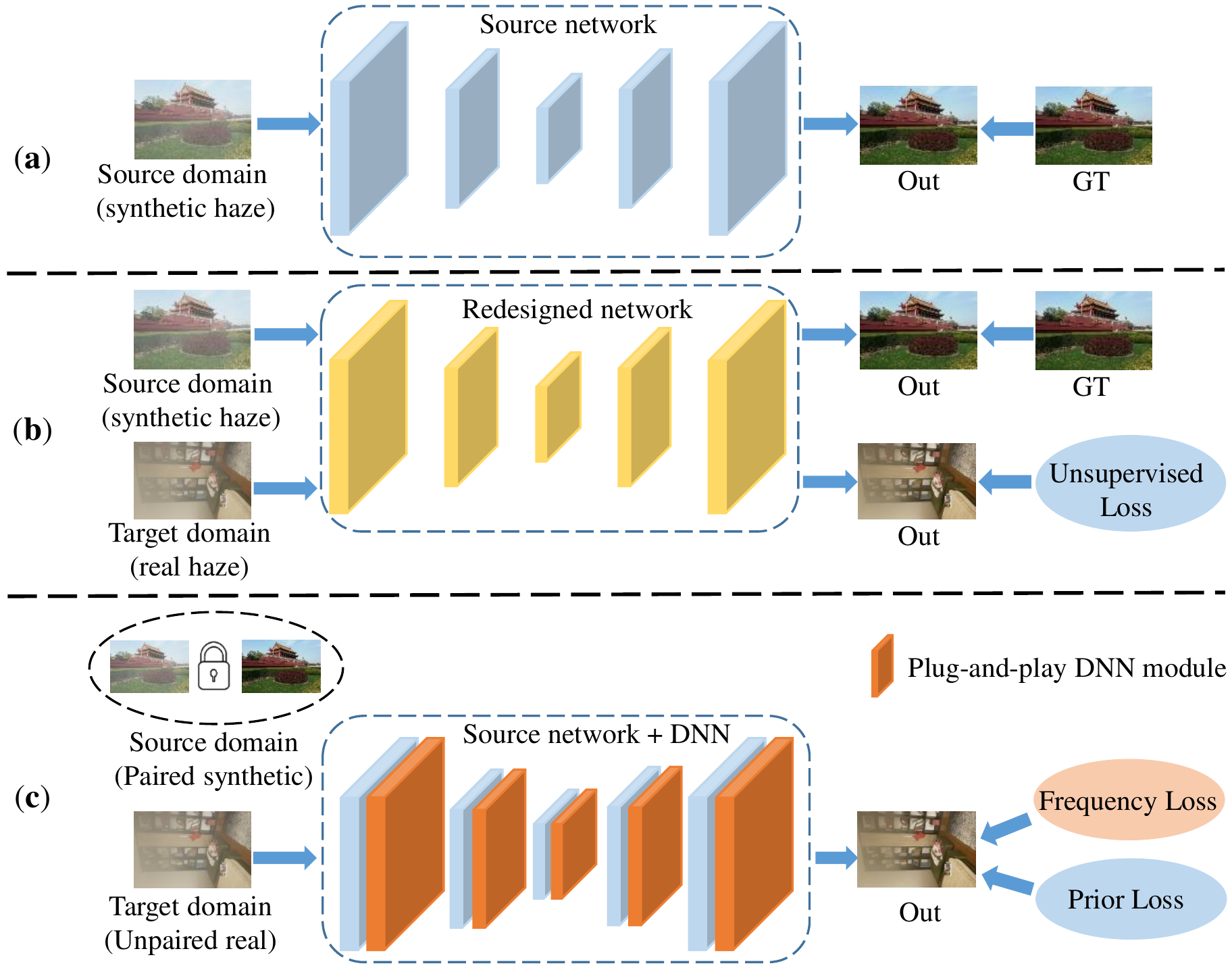}
  	\setlength{\abovecaptionskip}{-0.2cm}
    \setlength{\belowcaptionskip}{-0.5cm}
  \caption{(a) Source dehazing models that are trained on synthetic data (\emph{e.g.} MSBDN \cite{dong2020multi}). (b) Existing source-driven DA dehazing methods. These methods design customized architecture and need both source synthetic and target real hazy image during domain adaptation. (c) Our Source-Free Unsupervised Domain Adaptation image dehazing paradigm. Our method works with only target unpaired real hazy images available and can direct utilize well trained source models with frozen parameters in a plug-and-play manner.}
  \label{fig:demo}
\end{figure}

\noindent Haze is a common atmospheric phenomenon. Images captured in hazy environments usually suffer from noticeable visual quality degradation in object appearance and contrast, resulting in accuracy decreasing for subsequent visual analysis. Thus, image dehazing has been a focus of research in the computational photography and vision communities throughout the last decades.

As recognized, the hazing process can be represented by the physical scattering model \cite{mccartney1976optics}, which is usually formulated as
\begin{equation}
	\small
		I(x)=J(x) t(x)+A(1-t(x)),
\end{equation}
where $I(x)$ and $J(x)$ denote the hazy image and the clean
image respectively, $A$ is the global atmospheric light, and $t(x)$ is the transmission map. 

However, estimating the clean image from a single hazy input is an ill-posed and challenging problem. Given a hazy image $I(x)$, conventional prior-based dehazing algorithms attempt to estimate $t(x)$ and $A$ by constraining the solution space using a variety of sharp image priors \cite{berman2016non,he2010single,fattal2014dehazing,fattal2008single} and then restore the image via the scattering model. However, these hand-crafted image priors are based on specific observations, which may not be reliable for estimating the transmission map in the physical scattering model.

Numerous convolutional neural networks (CNNs)-based systems have been developed to estimate transmission maps \cite{cai2016dehazenet, ren2016single, zhang2018densely} or yield clean images directly \cite{li2017aod,li2018single,ren2018gated,liu2019griddehazenet,qu2019enhanced,qin2020ffa,dong2020multi,wu2021contrastive}, which achieve superior image dehazing performance over classic prior-based algorithms. However, these approaches require large quantities of paired hazy/clean images to train in a supervised learning manner. In general, due to the impracticality of acquiring large amounts of hazy-clean pairs in the real world, most dehazing models are trained on hazy synthetic datasets. In this paper, we denote these synthetic hazy dataset trained models as \textbf{source models}. However, source models often suffer from degraded performance on real-world hazy images due to the domain gap.

Recently, the above domain shift problem has drawn the attention of the image dehazing community. Existing Domain-Adaptation (DA) dehazing methods \cite{li2019semi,shao2020domain, ye2022mutual, li2020zero, li2021you, chen2021psd, liu2021synthetic} achieve remarkable performance, but they have two common drawbacks. First, they inevitably require full access to source synthetic hazy datasets to reduce the gap between the source synthetic and target real domains during model adaptation, which is cumbersome in practice due to storage, privacy, and transmission constraints. 
Second, most of these methods need to retrain source models with modification or use customized architecture for domain adaptation, neglecting to direct utilize numerous source models, which reduces the generality of their methods. 
In this paper, we denote these DA dehazing methods as \textbf{source-driven models}.

To address these problems, we present a novel Source-Free Unsupervised Domain Adaptation (SFUDA) image dehazing paradigm, where only a well-trained source model and an unlabeled target real hazy dataset are available during model adaptation. The overview of our method is shown in Fig. \ref{fig:demo}. Since the representations of features across synthetic and real domains vary significantly, our SFUDA achieves domain adaptation from the domain representation perspective. 
Specifically, the DRN module is devised for making the representation of real hazy image's features match that of the synthetic domain to adapt the source network. To implement this, the DRN module disentangles the features of real hazy images into domain-invariant and domain-variant parts. The domain-invariant part is obtained by Instance Normalization (IN), while the domain-variant part is implicitly guided by the feature statistics to make the features adapt to the source model. With IN equipped in the feature space, it can normalize feature statistics for style normalization \cite{huang2017arbitrary}. The proposed DRN module can be directly applied to existing dehazing models in a plug-and-play fashion. 

Additionally, the unsupervised frequency losses and physical prior losses are applied to regularize the dehazing processing of unlabeled real hazy images. 
Specifically, we explore the frequency property in our unsupervised setting. It is worth noting that although source network fails to remove the haze of real images, it well reconstructs the structural information. Therefore, we introduce the phase structure loss between the intermediate stages and output of the source network and the student network for structural consistency.
Besides, we find that the original real hazy image has an enhanced illumination and color contrast after being processed by Contrast Limited Adaptive Histogram Equalization (CLAHE). With this property, we introduce the amplitude style loss to regularize the enhanced illumination contrast between the output of CLAHE and the output of the student network.
Moreover, we employ the classical Dark Channel Prior (DCP) and Color Attenuation Prior (CAP) to form prior loss to exploit the inherent statistic property of haze-free images.

Existing source models that are equipped with our DRN module and unsupervised losses gain generalization power and can dehaze unlabeled  real hazy images. In conclusion, the main contributions of this work are summarized as follows:
\begin{itemize}
    \item We propose a novel Source-Free Unsupervised Domain Adaptation (SFUDA) image dehazing paradigm. To the best of our knowledge, this is the first attempt to address the problem of source-free UDA for image dehazing. Our method can be directly applied to existing dehazing models in a plug-and-play fashion, which is more general in practical. 
	\item we propose the Domain Representation Normalization (DRN) module to regularize representations of features in the real hazy domain to adapt to the frozen source network. 
	\item To train the whole framework in an unsupervised manner, we leverage the frequency property and physical priors to constrain the adaptation and generate more natural images.
	\item Extensive experiments on multiple baselines demonstrate the validity of our method visually and quantitatively. In particular, our method outperforms the state-of-the-art source-driven UDA approaches under a source-free setting.
\end{itemize}

\section{Related work}

\begin{figure*}
	\begin{center}
		\includegraphics[width=1.0\linewidth]{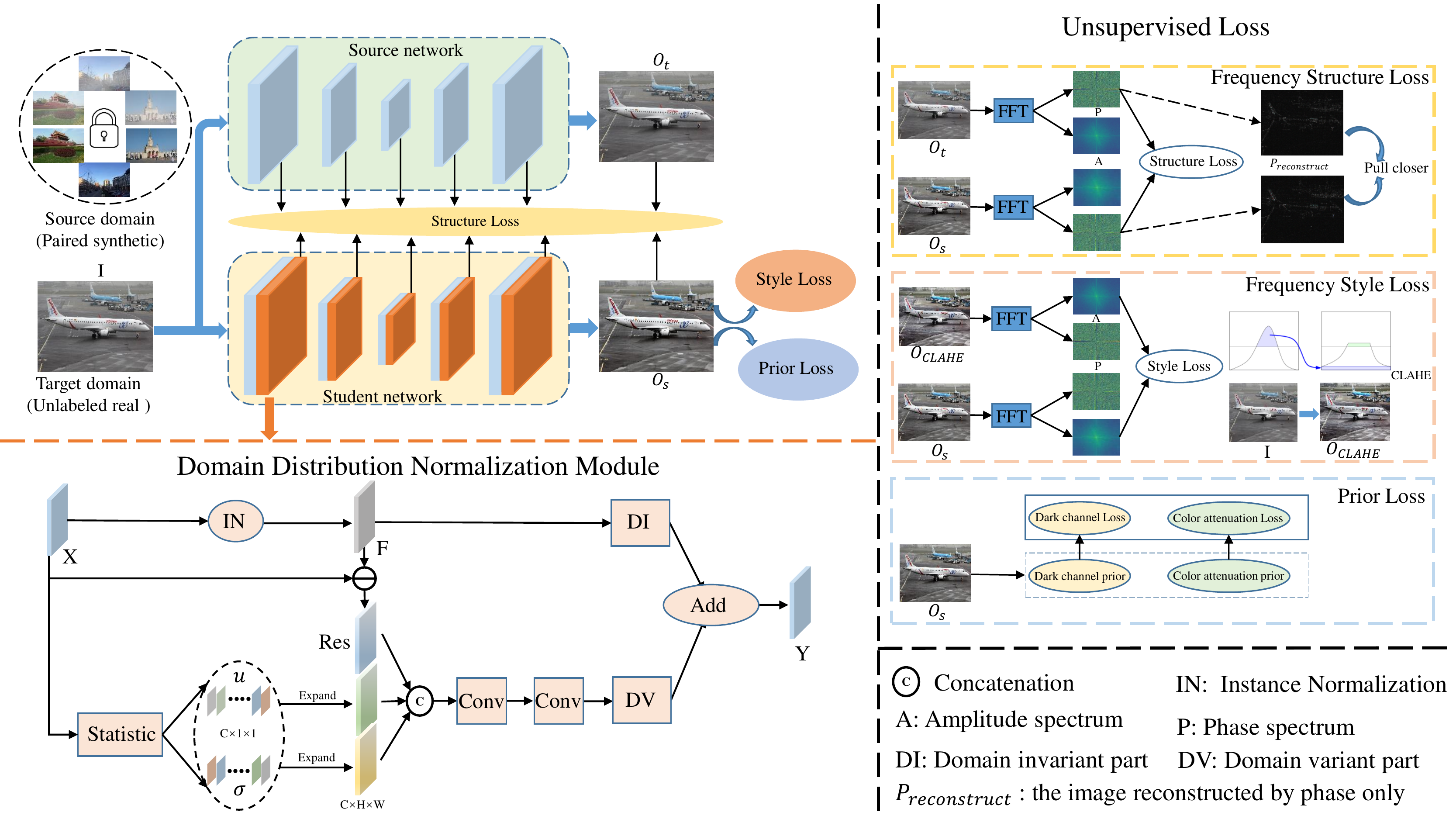}
	\end{center}
	\vspace{-3mm}
	\caption{Overview of the proposed framework. The source network can be any synthetic data pretrained model (e.g., MSBDN), while the student network is built by inserting our DRN module into the intermediate stages of the source model. Note that the parameters in the teacher network and student network’s source model part are frozen.}
	\vspace{-3mm}
	\label{fig:pipeline}
\end{figure*}

\subsection{Image dehazing}
\noindent Recent years have witnessed significant advances in image processing~\cite{Yao2019spectral,pan2022towards,wang2021jpeg}, including single image dehazing. Existing image dehazing methods can be roughly categorized into physical-based methods and deep learning-based methods.

\textbf{Physical-based.} Physical-based methods depend on the physical model~\cite{mccartney1976optics} and the handcraft priors from empirical observation, such as dark channel prior~\cite{he2010single}, color line prior~\cite{fattal2014dehazing}, and sparse gradient prior~\cite{chen2016robust}. 
However, the density of haze can be affected by various factors, which makes the haze formation at individual spatial locations space-variant. Therefore, the haze usually cannot be accurately characterized by merely a single transmission map. 

\textbf{Deep learning-based.} Different from the physical-based methods, deep learning-based methods employ convolution neural networks and large-scale datasets to learn the image prior \cite{cai2016dehazenet,li2017aod,ren2016single,zhang2018densely,liu2018learning,liu2019learning} or directly learn hazy-to-clear image translation \cite{ren2018gated,liu2019griddehazenet,qu2019enhanced,dong2020physics,deng2020hardgan,hong2020distilling,dong2020multi,qin2020ffa,wu2021contrastive,ye2021perceiving,wang2021eaa,zheng2021ultra}.
For example, MSBDN~\cite{dong2020multi} proposes a boosted decoder to progressively restore the haze-free images. 
Existing methods have shown outstanding performances on image dehazing. 
However, they train on paired synthetic data and generalize poorly on real-world data.

\subsection{Domain adaptation dehazing}
\noindent Recently, some domain adaptation dehazing methods \cite{li2019semi,shao2020domain, ye2022mutual, shyam2021towards, li2020zero,li2021you, chen2021psd, liu2021synthetic} have been proposed to tackle domain shift between synthetic and real domains. For example, Shao et al. \cite{shao2020domain} developed a domain adaptation paradigm, which consists of a image translation module and two dehazing modules and the image translation module is used for data augmentation.
Chen et al. \cite{chen2021psd} modified and retrained an existing dehazing model with physical priors and fine-tune it on both synthetic and real hazy images. 



However, these existing techniques require the full access to source synthetic hazy datasets during model adaptation, which limits their practical application, due to the non-availability of source datasets in some cases.

\subsection{Source-Free Unsupervised Domain Adaptation}
\noindent Since labeled source data may not be available in some real-world scenarios due to data privacy issues, Source-Free Unsupervised Domain Adaptation aims to explore how to improve performance of an existing source model on the target domain with only unlabeled target data available.
Recently, this new domain adaptation paradigm has been applied to different tasks \cite{liu2021source, li2020model, chidlovskii2016domain, huang2021model, ahmed2021unsupervised} for its high practical value and easy-to-use property.
Different from these techniques, we first attempt to introduce SFUDA to image dehazing tasks, and innovatively address this problem from the domain representation perspective. Our solution is general and can be directly applied to existing source dehazing models.

\section{Method}

\subsection{Method Overview}
\noindent 
Source dehazing models are trained on synthetic data, thus obtaining the dehazing knowledge for the representation of synthetic domain. With this property in mind, we design the DRN module from the domain representation perspective to adapt to the source model. Besides, we design the unsupervised loss to constrain the learning process. Our method makes the first attempt to directly leverage the learned dehazing knowledge of existing source models, providing a general solution for domain adaptation and works in a source-free and plug-and-play manner.


As shown in Fig. \ref{fig:pipeline}, our domain adaptation framework works in a teacher-student mode. The teacher network can be any existing source dehazing models (\emph{e.g.} FFA-Net \cite{qin2020ffa}, MSBDN \cite{dong2020multi}, we choose MSBDN as the source model by default), while the student network consists of our DRN module and the source model. DRN module is expected to make the representation of real data match the synthetic domain to adapt the frozen source network.
Besides, although the source network performs poorly on real hazy images, it owns the ability to preserve the structure of images. Thus, we froze the parameter of the source network during training so that the structure information of the source network can provide supervision for the student network. The parameter of the student network has two parts: the DRN module part and the source network part. Only the parameter of the DRN module is updated during back-propagation to keep the dehazing ability of the source model on synthetic domain representation. 

\begin{figure}[thbp]
  \centering
  \includegraphics[width=\linewidth]{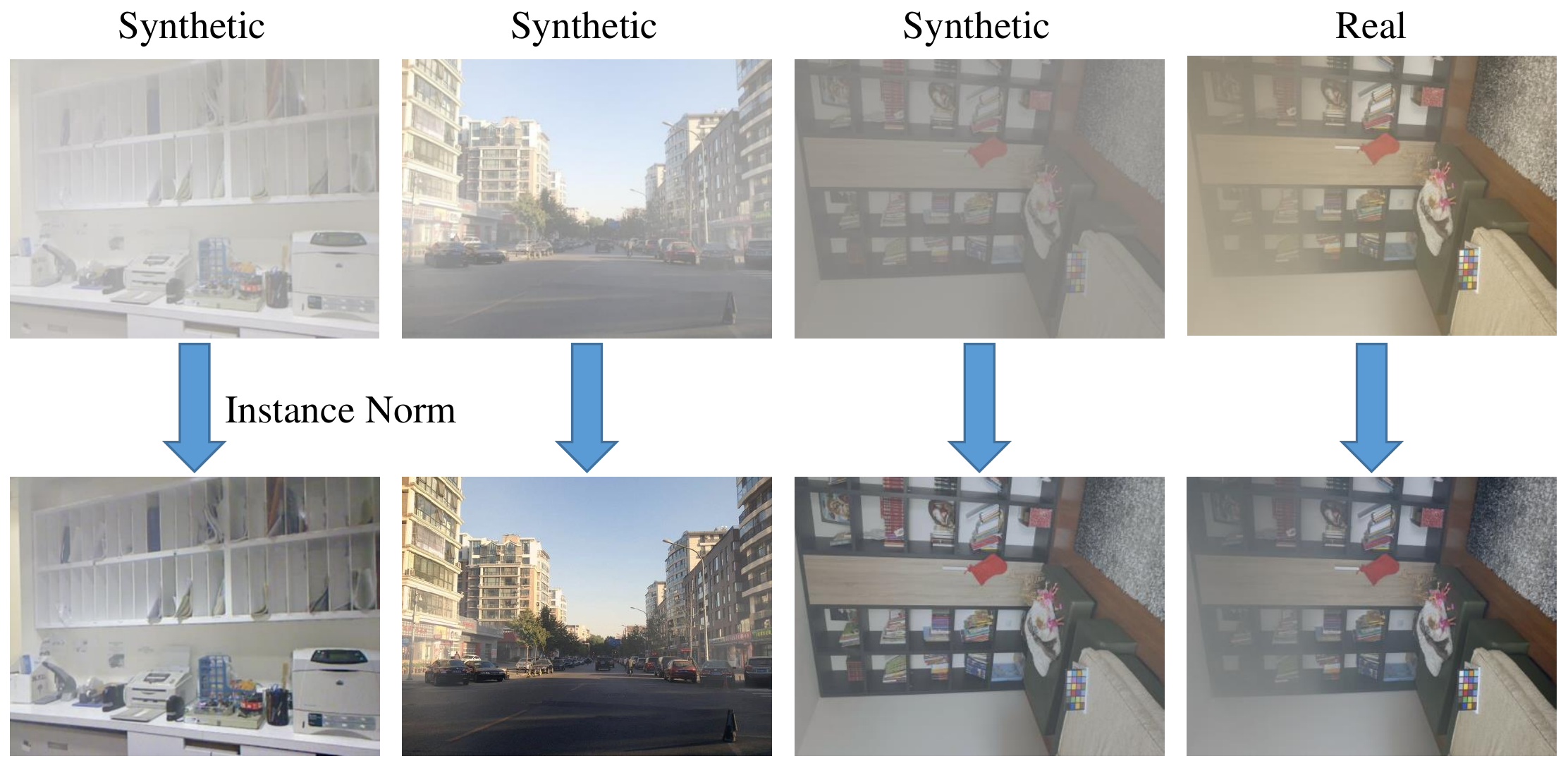}
  	\setlength{\abovecaptionskip}{-0.2cm}
    \setlength{\belowcaptionskip}{-0.5cm}
  \caption{Examples of both real and synthetic hazy images after Instance Normalization operation.}
  \Description{IN}
  \label{fig:IN}
\end{figure}

\subsection{Domain Representation Normalization Module}
Existing normalization-based methods \cite{jia2019instance, fan2021adversarially} regularize the distribution by normalize it to the standard normal distribution and then modulate it to another distribution with learned statistics. After such $\sigma_{1}\frac{X - \mu}{\sigma }  + \mu_{1}$ operation, the distribution change from original $N\left ( \mu,\sigma^{2} \right )$ to learned $N\left ( \mu_{1},\sigma_{1}^{2} \right )$, where $N\left ( \mu,\sigma^{2} \right )$ denotes normal distribution with mean $\mu$ and standard deviation $\sigma$. However, although such a strategy works well for high-level tasks, normalization will inevitably induce the loss of the image discriminative features \cite{jin2020style, pan2018two} for image reconstruction.

Differently, for input real hazy domain feature $X(x)\in \mathbb{R}^{C \times H \times W}$, where $C$, $H$ and $W$ represent channel dimension, height, and width, respectively, our DRN module disentangles it into domain invariant knowledge $DI$ and domain variant knowledge $DV$. We introduce Instance normalization (IN) to get $DI$. IN performs style normalization by normalizing feature statistics, which have been found to carry the style information of an image \cite{huang2017arbitrary}.

We show some hazy images after IN operation in Fig. \ref{fig:IN}. The synthetic and real hazy images of the same scene are pulled closer in style after IN, which demonstrates that IN property of excavating domain-invariant features fits image dehazing task. 

\textbf{For the domain invariant part}, $X(x)$ is first processed by IN to get $F(x)$, formulated by:
\begin{equation}
	\small
		\begin{aligned}
        {F}(x) =Instance Norm(X) =  \frac{X - \mu}{\sigma },
        \end{aligned}
\end{equation}
where, $\mu\in \mathbb{R}^{C \times 1 \times 1}$ and $\sigma\in \mathbb{R}^{C \times 1 \times 1}$ denote mean and standard deviation of $X(x)$ along the channel dimension. $F(x)$ represents domain-invariant features of standard normal distribution.

\textbf{For the domain variant part}, the residual between input $X(x)$ and domain invariant part $F(x)$ are denoted as $Res(x)$, which can be regarded as domain variant part. Besides, we also obtain the mean and standard deviation of $X(x)$.
\begin{equation}
	\small
        \mu=\frac{1}{m} \sum_{k \in S_{i}} x_{k}, \quad \sigma=\sqrt{\frac{1}{m} \sum_{k \in S_{i}}\left(x_{k}-\mu_{i}\right)^{2}+\epsilon}
\end{equation}
with $\epsilon$ be a small constant for numerical stability. Unlike previous methods, we implicitly incorporate $\mu$ and $\sigma$. Concretely, we first expand $\mu$ and $\sigma$ along the spatial dimension to get the same size as $Res(x)$. Afterwards, the expanded $\mu$, $\sigma$ and $Res(x)$ are concatenated along the channel dimension to implicitly guide the learning of $Res(x)$. The final domain variant knowledge $DV$ can be computed as:
\begin{equation}
	\small
        DV = Conv(Cat\left ( Res , \mu ,\sigma  \right )).
\end{equation}

\begin{figure}[thbp]
  \centering
  \includegraphics[width=\linewidth]{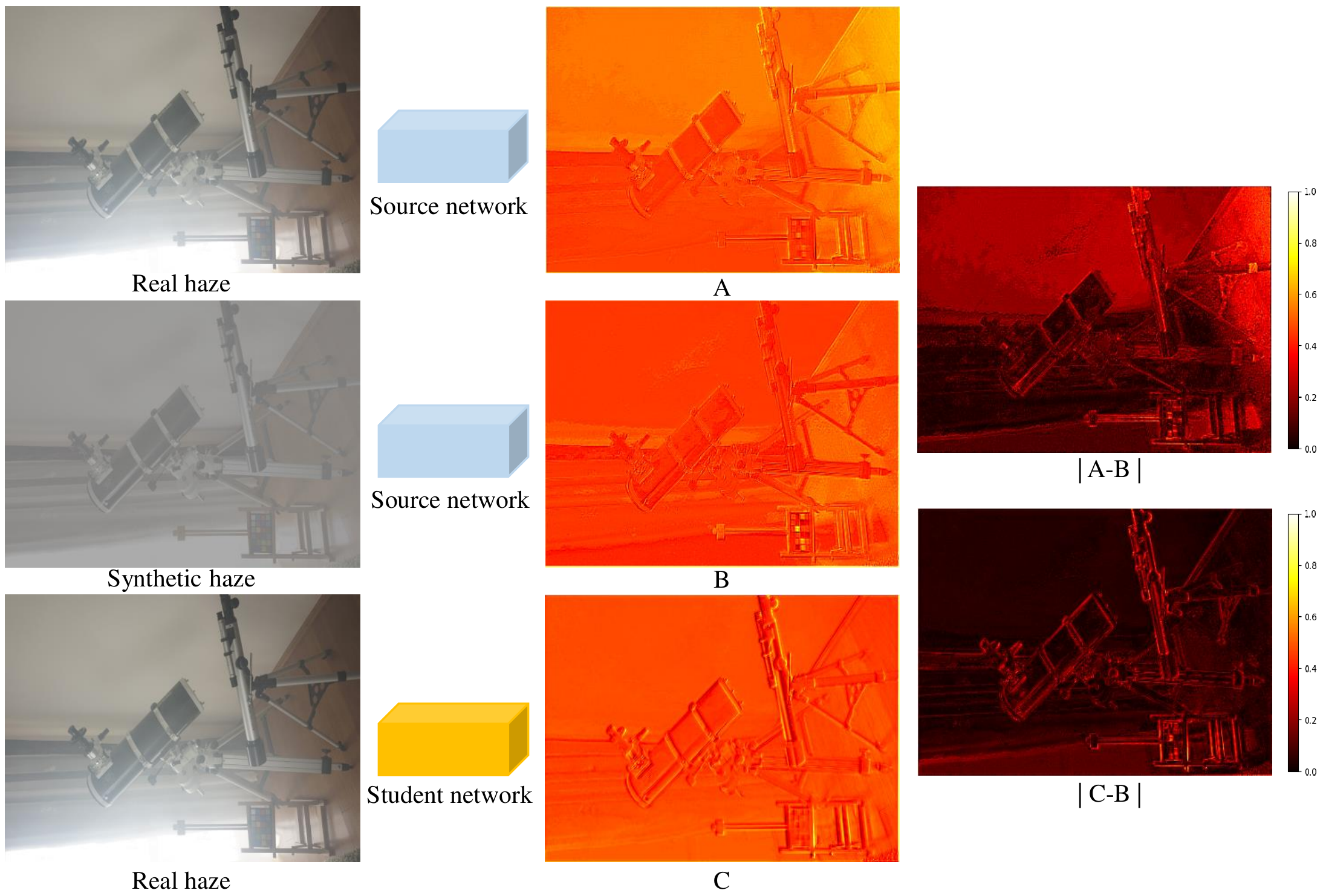}
  	\setlength{\abovecaptionskip}{-0.2cm}
    \setlength{\belowcaptionskip}{-0.5cm}
  \caption{Feature visualization on the effectiveness of our DRN module. The second column is intermediate features.}
  \Description{demo}
  \label{fig:features}
\end{figure}

Finally, we integrate domain invariant knowledge $DI$ with domain variant knowledge $DV$ to get the final output $Y(x)$ of the DRN module. Our DRN module keeps domain invariant knowledge $DI$ and modifies the domain variant knowledge along with the loss constraints in Sec.~\ref{losses}, which make the representation of output $Y(x)$ match the synthetic representation. To better testify this, we show the feature visualization in Fig. \ref{fig:features}. It is evident that real hazy image features in our student network are more similar to the synthetic domain representation than real hazy image features in the source network.

\subsection{Training Losses}\label{losses}

\begin{figure}[htbp]
  \centering
  \includegraphics[width=\linewidth]{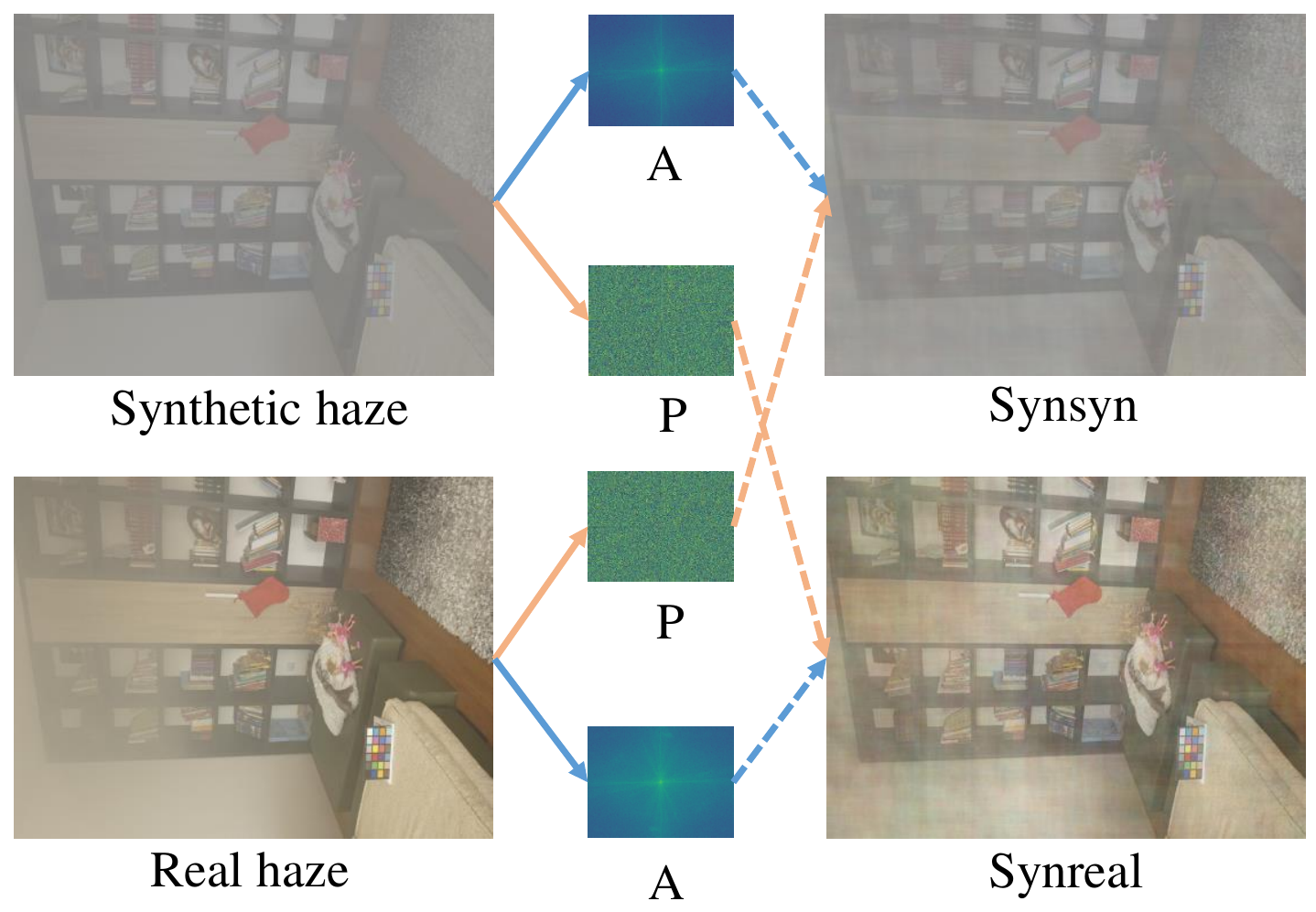}
  	\setlength{\abovecaptionskip}{-0.2cm}
    \setlength{\belowcaptionskip}{-0.5cm}
  \caption{Visualization on the relationship between the haze degradation and the characteristics of amplitude spectrum and phase spectrum in frequency domain.}
  \Description{exchange}
  \label{fig:exchange}
\end{figure}

Our framework works in an unsupervised way. Therefore, it is critical to design proper loss function as supervision to drive the DRN module learning. Specifically, We explore the frequency property in our unsupervised setting and innovatively introduce two frequency losses for keeping domain-invariant structure and modulating domain-variant style. Besides, we select two effective and well-grounded physical priors to provide us with the prior knowledge of real images.

\subsubsection{\textbf{Frequency domain Losses}}
As is known, the illumination contrast of an image is represented by the amplitude spectrum, while the structure information is represented by the
phase spectrum \cite{oppenheim1981importance,skarbnik2009importance}. Based on this theory, we further conduct an amplitude and phase exchange experiment to validate its applicability in our setting. As shown in Fig. \ref{fig:exchange}, we exchange the amplitude and phase of synthetic and real hazy images. The results demonstrate that synthetic and real hazy images of the same scene have approximately the same phase (structure) spectrum, while the style and haze degradation mainly manifests in the amplitude spectrum.

\textbf{Phase (Structure) Loss.}
The source dehazing network can not generalize well to real hazy images, but they can still retain satisfactory structural information, which is an important dehazing prior to unsupervised learning. With this prior knowledge in mind, we design a Phase (Structure) Loss.

Concretely, we explore the information contained in the phase spectrum of both intermediate features and output images from teacher to student to provide structural constraints. The outputs of the teacher and student network are denoted as $O_{t}$ and $O_{s}$. We first transform the output image to the frequency domain by Fourier Transform.
\begin{equation}
	\small
	\label{FFT}
	{F}(x)(u, v)=\sum_{h=0}^{H-1} \sum_{w=0}^{W-1} x(h, w) e^{-j 2 \pi\left(\frac{h}{H} u+\frac{w}{W} v\right)} .
\end{equation}
The frequency-domain feature $ F(x)$ is denoted as $ F(x) =  R(x) + j I(x)$, where $ R(x)$ and $ I(x)$ represent the real and imaginary part of $ F(x)$. Then the real and imaginary parts are converted to amplitude and phase spectrums, which can be formulated as:
\begin{equation}
	\small
	\begin{array}{l}
	\label{realtoamp}
    {A}(x)(u, v)=\left[ R^{2}(x)(u, v)+  I^{2}(x)(u, v)\right]^{1 / 2}, \\
    {P}(x)(u, v)=\arctan \left[\frac{ I(x)(u, v)}{ R(x)(u, v)}\right],
    \end{array}
\end{equation}
where $ A(x)$ is the amplitude spectrum, $ P(x)$ is the phase spectrum. The phase spectrums of teacher and student network outputs are denoted as $ P_{t}(x)$ and $ P_{s}(x)$. Then the Phase (Structure) Loss are formulated as:
\begin{equation}
	\small
	\label{Structure}
		{L}_{Pha}=\frac{2}{U V} \sum_{u=0}^{U / 2-1} \sum_{v=0}^{V-1}    \left\|\left|P_{t}\right|_{u, v}-|P_{s}|_{u, v}\right\|_{1} .
\end{equation}
Note, in our implementation, the summation for $u$ is only performed up to $U/2-1$, since $50\%$ of all frequency components are redundant. Moreover, this phase loss is also applied to intermediate features in the same way.

\textbf{Amplitude (Style) Loss.}
Contrast Limited Adaptive Histogram Equalization (CLAHE) \cite{pizer1987adaptive, reza2004realization} is a classical digital image processing technology for improving image contrast and it is also applied to image dehazing.
However, it is inadvisable to directly apply CLAHE-enhanced images as supervision since CLAHE inevitably introduces bottom noise and brings inherent flaws into our network. To address this problem, we leverage the mentioned frequency property that the illumination contrast of an image is represented by the amplitude spectrum and design the Amplitude (Style) Loss to utilize CLAHE-enhanced illumination and contrast.

Specifically, the amplitude spectrum of student network output $O_{s}(x)$ is denoted as $ A_{s}(x)$ and the amplitude spectrum of CLAHE-enhanced real hazy image is denoted as $ A_{CLAHE}(x)$. Then the Amplitude (Style) Loss is formulated as:
\begin{equation}
	\small
		{L}_{Amp}=\frac{2}{U V} \sum_{u=0}^{U / 2-1} \sum_{v=0}^{V-1}    \left\|\left|A_{s}\right|_{u, v}-|A_{CLAHE}|_{u, v}\right\|_{1} .
\end{equation}

\subsubsection{\textbf{Physical Prior Losses}}
Various handcraft priors, such as dark channel prior~\cite{he2010single}, color line prior~\cite{fattal2014dehazing}, color attenuation prior~\cite{zhu2014single}, sparse gradient prior~\cite{chen2016robust}, maximum reflectance prior~\cite{zhang2017fast}, and non-local prior~\cite{berman2016non}, are drawn from empirical observation and reflect inherent statistic property of haze-free images. 
From these priors, we select two effective and well-grounded ones to provide us with the prior knowledge of real images. Additionally, unlike \cite{chen2021psd, liu2021synthetic}, 
we employ these prior losses with only output images available.

\textbf{Dark Channel Prior (DCP) Loss.}
Dark Channel Prior (DCP) \cite{he2010single} is the most famous and effective prior for image dehazing. DCP is a statistical property of outdoor haze-free images: most patches in these images should contain pixels that are dark in at least one color channel. We formulate DCP as a loss function as in \cite{li2020zero}.
\begin{equation}
	\small
	\label{DCP}
		{L}_{DCP}=\left\|\min _{c \in\{r, g, b\}}\left(F^{c}(y)\right)\right\|_{p} ,
\end{equation}
where $F^{c}(\cdot)$ is the c-th color channel of $y$, and $y$ is a local patch of the student network output $O_{s}(x)$. With this dark channel loss, our student network incorporates the statistical properties from the recovered clean images, avoiding an explicit ground truth on the recovered image.

\textbf{Color attenuation prior (CAP) Loss.}
Color attenuation prior (CAP) \cite{zhu2014single, zhu2015fast} can be explained as,
in a haze-free region, the difference between the brightness and the saturation is close to zero. In concrete implementation, the difference between the value and saturation in the predicted $O_{s}(x)$ should be minimized as small as possible. We formulate CAP as a loss function as in \cite{li2021you}.
\begin{equation}
	\small
	\label{CAP}
		{L}_{CAP}=\left\|V\left(O_{s}(x)\right)-S\left(O_{s}(x)\right)\right\|_{p} ,
\end{equation}
where $V\left(O_{s}(x)\right)$ and $S\left(O_{s}(x)\right)$ respectively denotes the brightness and saturation of $O_{s}(x)$.

\textbf{Total Loss.}
The final loss function used in our method can be defined as:
\begin{equation}
	\small
		{L}=\lambda_{p} {L}_{Pha} + \lambda_{a} {L}_{Amp} + \lambda_{d} {L}_{DCP} + \lambda_{c} {L}_{CAP},
\end{equation}
where $\lambda _{p}$, $\lambda _{a}$, $\lambda _{d}$ and $\lambda _{c}$ are weight factors.


\begin{figure*}
	\begin{center}
		\includegraphics[width=1.0\linewidth]{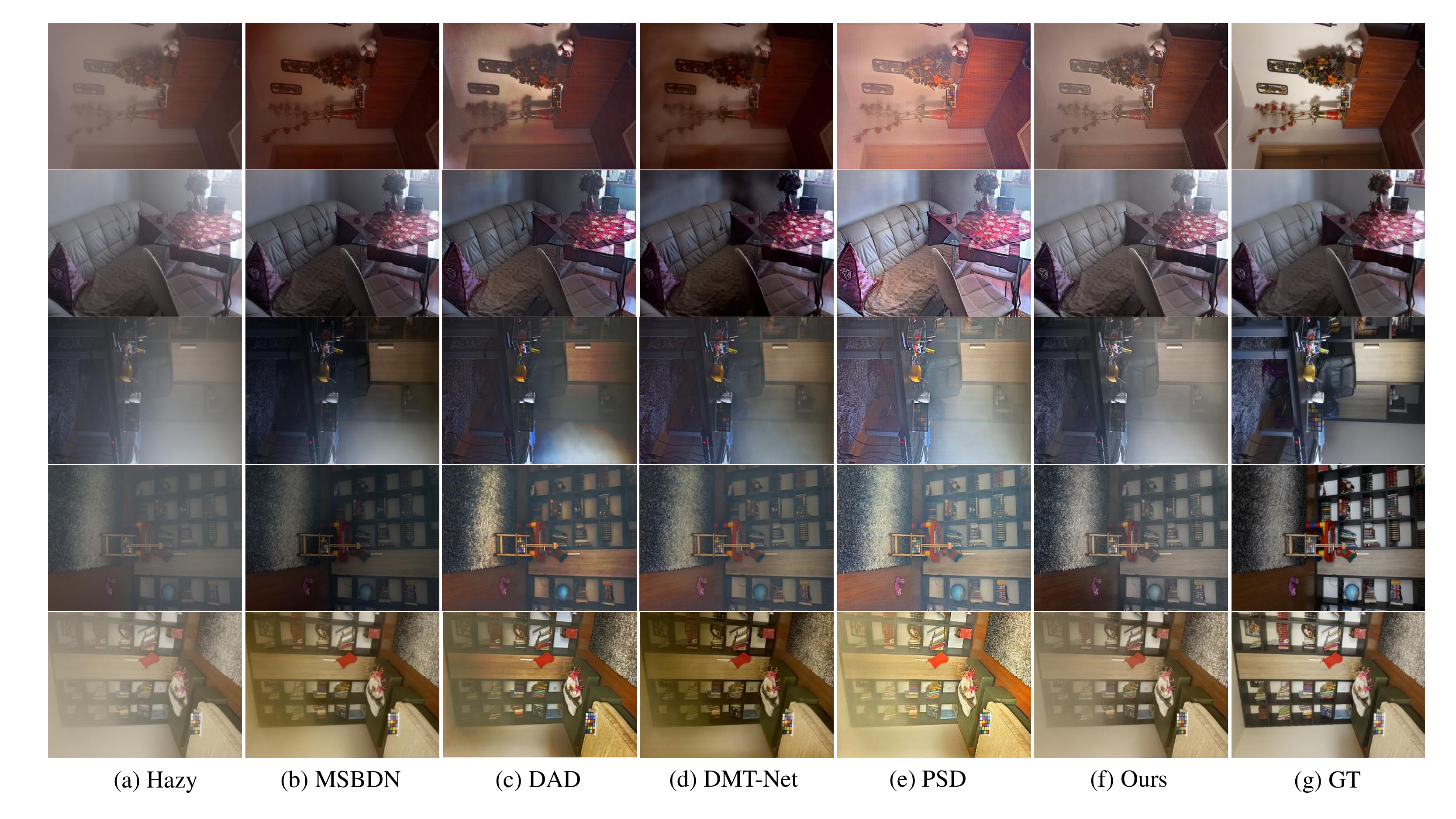}
	\end{center}
	\vspace{-3mm}
	\caption{Visual comparison of source baseline and DA dehazing results on labeled real hazy dataset I-Haze.}
	\vspace{-3mm}
	\label{fig:I-Haze}
\end{figure*}

\section{Experiments}


\subsection{Experiment Setup}
\noindent \setlength{\parindent}{2em}\textbf{Datasets.}
We select Unannotated Real Hazy Images (URHI) from RESIDE dataset \cite{li2018benchmarking} for domain adaptation training. For evaluating the effectiveness of our method, we test on two real-world datasets, RTTS \cite{li2018benchmarking} and I-Haze \cite{ancuti2018haze}. RTTS is a subset of RESIDE dataset \cite{li2018benchmarking}, consisting of $4322$ real unlabeled outdoor hazy images. I-Haze contains 30 paired real indoor hazy images.

\textbf{Metrics.}
We employ two widely used metrics, the Peak Signal to Noise Ratio (PSNR) and the Structural Similarity Index (SSIM), to quantitatively assess the results on the labeled I-Haze dataset. Besides, we utilize two well-known no-reference image quality assessment indicators: BRISQUE \cite{mittal2012no} and NIQE  \cite{mittal2012making} to assess the results on the unlabeled RTTS dataset.

\begin{figure*}
	\begin{center}
		\includegraphics[width=1.0\linewidth]{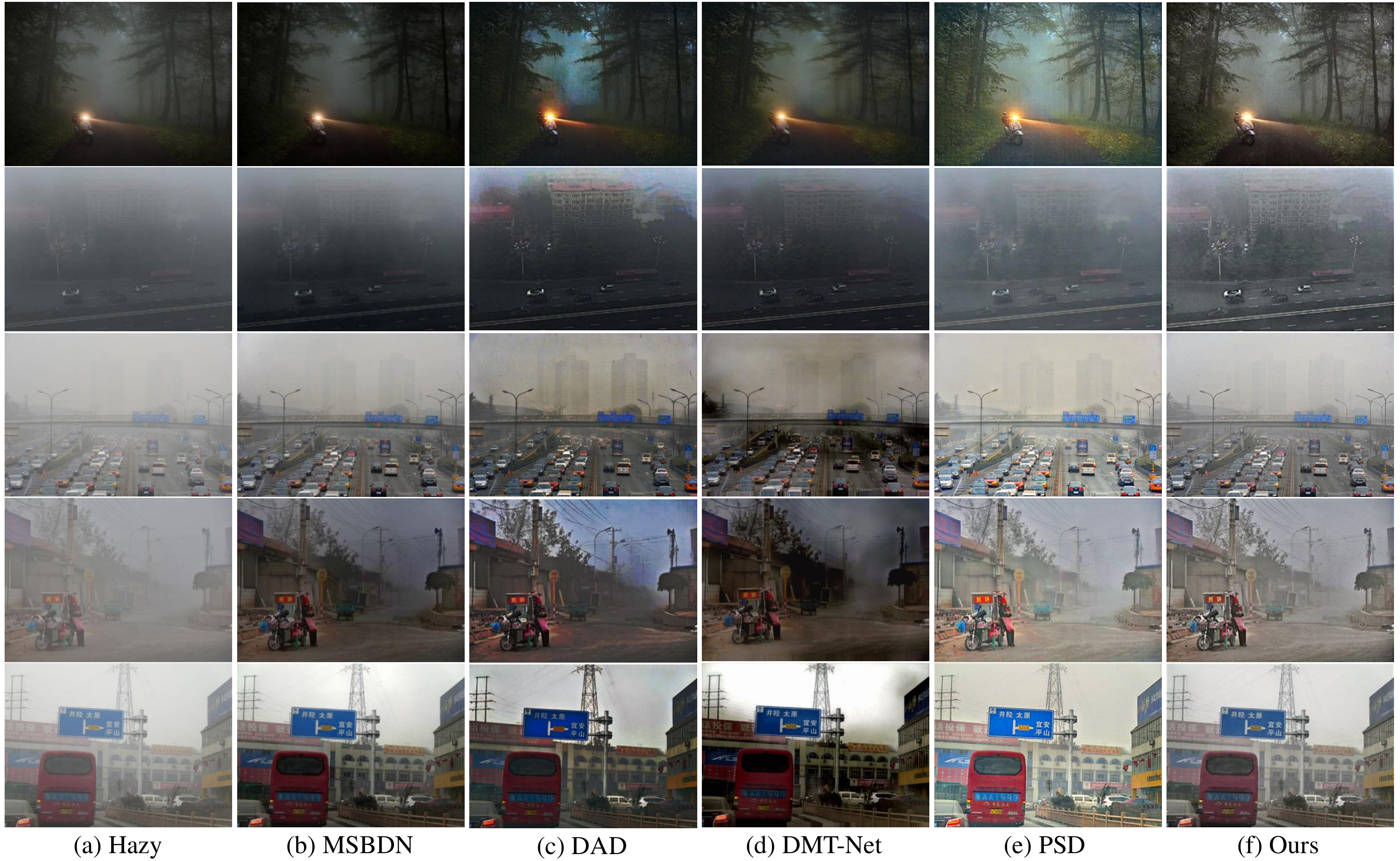}
	\end{center}
	\vspace{-3mm}
	\caption{Visual comparison of DA dehazing results on unlabeled real hazy dataset RTTS.}
	\vspace{-3mm}
	\label{fig:RTTS}
\end{figure*}

\textbf{Implementation Details.} 
We use ADAM as the optimizer with $\beta_{1}=0.9$, and $\beta_{2}=0.999$, and the initial learning rate is set to $1\times10^{-4}$. 
The learning rate is adjusted by the cosine annealing strategy~\cite{he2019bag}. The training epoch, batch and patch sizes are set to $10$, $6$, and $256\times256$, respectively. The trade-off weights in loss function are set to $\lambda _{p} = 1$, $\lambda _{a} = 1$, $\lambda _{d} = 10^{-3}$, and $\lambda _{c} = 10^{-3}$.

\begin{table}[htbp]
	\caption{Quantitative comparisons with state-of-the-art (SOTA) DA methods on two real-world dehazing datasets. $\downarrow$ denotes lower is better, while $\uparrow$ means higher is better. }
\label{Performance}
\centering
\begin{tabular}{c|c|cc|cc}
	\hline 
	\multirow{2}{*}{} & \multirow{2}{*}{Method} & \multicolumn{2}{c|}{I-HAZE \cite{ancuti2018haze}} & \multicolumn{2}{c}{RTTS \cite{li2018benchmarking}} \\ \cline{3-6}
	&  & PSNR$\uparrow$ & SSIM$\uparrow$  & BRISQUE$\downarrow$  & NIQE$\downarrow$ \\ \hline 
	& Hazy & - & - & 36.703  & 5.209  \\
	\hline 
	Baseline & MSBDN \cite{dong2020multi}  & 16.623  & 0.780 & 32.575  & 4.865    \\  \hline
	DA & DAD \cite{shao2020domain}   & 12.076  & 0.750   & 34.423  & 5.367    \\
	DA & DMT \cite{liu2021synthetic} & 12.902  & 0.521   & 31.594  & 4.963    \\
	DA & PSD \cite{chen2021psd}      &14.575   &0.781    & 28.011  & 4.494   \\ 
	\hline
	DA & Ours  & \textbf{17.602} &\textbf{0.802} &\textbf{27.330} & \textbf{4.326} \\ 
	\hline
\end{tabular}
\end{table}

\subsection{Comparison with State-of-the-art Methods}

\noindent We compare the performance of our SFUDA with three SOTA DA dehazing methods: DAD \cite{shao2020domain}, PSD \cite{chen2021psd}, and DMT-Net \cite{liu2021synthetic}. Amounts of experiments are conducted on two real-world hazy datasets. 

\subsubsection{\textbf{Comparison on labeled I-Haze Dataset}}
\textbf{Visual Quality.}
As shown in Fig. \ref{fig:I-Haze}, Compared with the ground truths, it is evident that the results of MSBDN, DAD, and DMT not only fail to remove the dense haze but also suffer from color shift. Moreover, the severe color distortion problem of PSD is clearly exposed under ground-truth haze-free images.
Compared with all these methods, our method generates the highest-fidelity dehazed results. 

\textbf{Quantitatively results.}
Table \ref{Performance} compares the quantitative results of different methods on the I-Haze dataset, which indicates our source-free method achieves the best performance with 17.602dB PSNR and 0.802 SSIM. What's more, different from outdoor datasets URHI and RTTS, I-Haze is an indoor dataset and its representation is far away from that of the training dataset URHI than the RTTS dataset, which is shown in the supplementary material. Consequently, our best performance on the I-Haze dataset strongly proves the generalization ability of our method.

\subsubsection{\textbf{Comparison on unlabeled RTTS Datasets}}
\textbf{Visual Quality.}
As shown in Fig. \ref{fig:RTTS}, the source baseline model MSBDN remains haze residual, especially in distant areas. The dehazing results of DAD and DMT are dark in some regions. PSD tends to produce visually satisfactory images, but it over amplifies color contrast by direct use of CLAHE prior, thus seems to be unnatural. 
In contrast, our SFUDA generates high-quality haze-free images with more natural color, clearer architecture, and finer details.

\textbf{No-Reference Image Quality Assessment.}
For quantitative comparison of the unlabeled dataset, we employ two well-known no-reference image quality assessment indicators: BRISQUE \cite{mittal2012no} and NIQE  \cite{mittal2012making}. As shown in table \ref{Performance}, our method largely outperforms baseline MSBDN. Besides, it also surpasses the source-driven DA methods DAD, DMT, and PSD.

\begin{figure*}
	\begin{center}
		\includegraphics[width=1.0\linewidth]{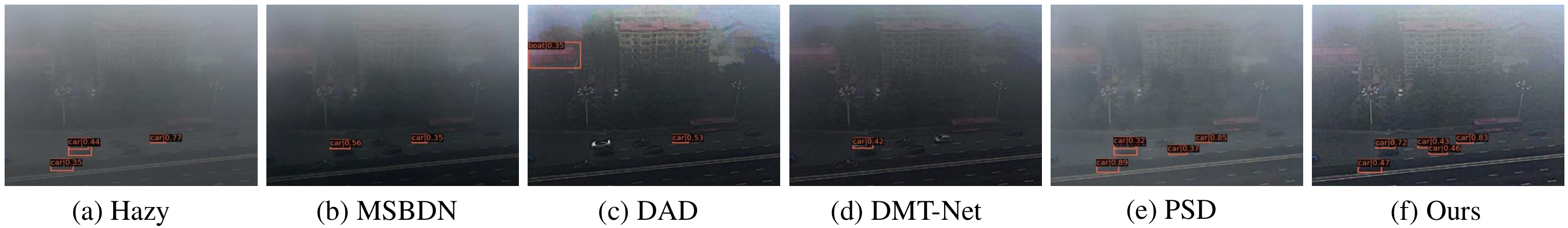}
	\end{center}
	\vspace{-3mm}
	\caption{Detection results of SOTA DA methods on an representative image sampled from RTTS dataset.}
	\vspace{-3mm}
	\label{fig:detection}
\end{figure*}

\textbf{Task-Driven Evaluation.}
The performance of high-level computer vision tasks such as object detection and scene understanding is considerably influenced by input images captured in hazy scenes \cite{li2019single, vidalmata2020bridging, yang2020advancing}. Thus, image dehazing task can be used as a preprocessing step for these high-level tasks. Inversely, the performance of these downstream tasks can also be used to evaluate the effectiveness of image dehazing algorithms. To this end, we testify our method and other SOTA dehazing methods on the RTTS dataset with object detection task. We show the detection result of a representative image sampled from RTTS in Figure \ref{fig:detection}. It is evident that the detection performance of our method surpasses the baseline MSBDN and DA dehazing methods DAD and DMT-Net. Besides, it achieves comparable results compared to PSD.

\begin{table}[htbp]
	\caption{Quantitative comparisons with baseline methods and their vanilla direct fine-tuning on two real-world dehazing datasets. ``ft'' represents fine-tuned baseline.}
\label{Baseline}
\centering
\begin{tabular}{c|cc|cc}
	\hline 
	\multirow{2}{*}{Method} & \multicolumn{2}{c|}{RTTS \cite{li2018benchmarking}} & \multicolumn{2}{c}{I-HAZE \cite{ancuti2018haze}}  \\ 
	\cline{2-5}
	         & BRISQUE$\downarrow$    &NIQE$\downarrow$   &PSNR$\uparrow$   &SSIM$\uparrow$ \\
	\hline
	FFA-Net         & 59.145        & 8.212         & 10.105         & 0.552  \\
	FFA-Net (ft)     & 42.145        & 7.393         & 12.151         & 0.638  \\
	Ours (FFA-Net)   &\textbf{29.311}&\textbf{6.191} &\textbf{13.404} &\textbf{0.718} \\
	\hline
	MSBDN          & 32.575         & 4.865         & 16.623         & 0.780  \\  
	MSBDN (ft)      & 30.210         & 4.659         & 17.180         & 0.781   \\ 
	Ours (MSBDN)    &\textbf{27.330} &\textbf{4.326} &\textbf{17.602} &\textbf{0.802} \\
	\hline
\end{tabular}
\end{table}

\begin{figure*}
	\begin{center}
		\includegraphics[width=1.0\linewidth]{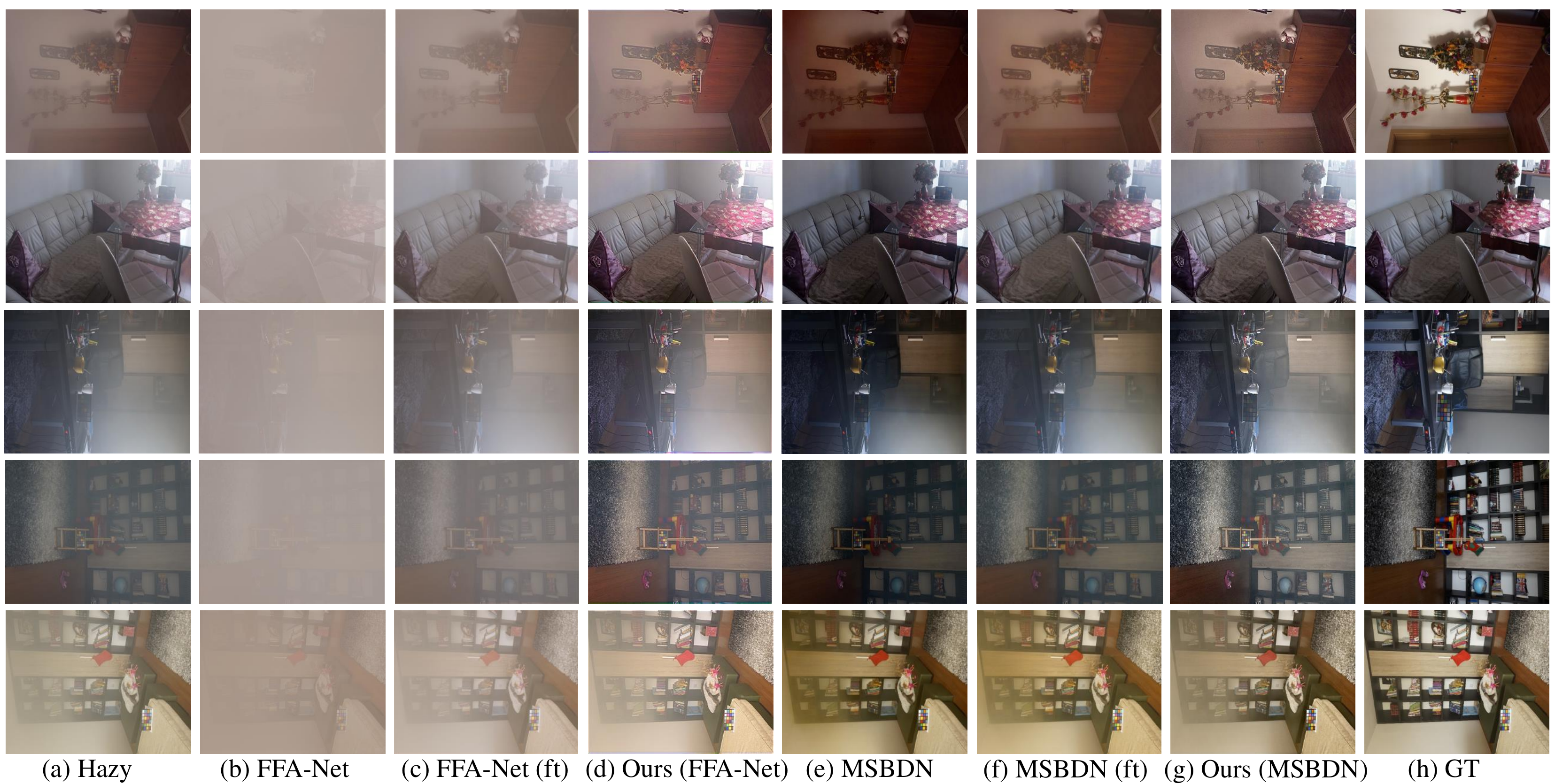}
	\end{center}
	\vspace{-3mm}
	\caption{Visual comparison with baselines and their fine-tuned (ft) models on labeled real hazy dataset I-Haze.}
	\vspace{-3mm}
	\label{fig:Baseline}
\end{figure*}

\subsection{Comparison with fine-tuned source models}
\noindent In order to compare more fairly with existing source baseline methods and further prove the superiority of our method, we compare with two state-of-the-art source baselines methods: FFA-Net \cite{qin2020ffa} and MSBDN \cite{dong2020multi} and their fine-tuned ones. Note that these source models are fine-tuned in the same setting as our method for fair comparison. 
The quantitative comparison results are presented in Table \ref{Baseline}. It is obvious that applying our method to these baseline methods is much more effective than vanilla direct fine-tuning. Not that our method improves performance with negligible additional parameters (MSBDN 31.35M, Ours(MSBDN) 31.93M).

Besides, we present the visual comparison in Fig. \ref{fig:Baseline}, vanilla direct fine-tuning ones still fails to remove the haze and suffer from color shift problem. In contrast, our method on these baselines generates natural and visually desirable results. 

\begin{table}[htbp]
\caption{Ablation study of our method on I-Haze dataset.}
\label{tab:ablation}
\begin{tabular}{c|ccccc|cc}
\hline
Label&DRN       & Str      & Sty      & DCP      & CAP       & PSNR (dB) & SSIM    \\ 
\hline
a    &$\times$  &\checkmark&\checkmark&\checkmark&\checkmark   & 17.180    & 0.781 \\
b    &\checkmark&$\times$  &\checkmark&\checkmark&\checkmark   & 17.124    & 0.781 \\
c    &\checkmark&\checkmark&$\times$  &\checkmark&\checkmark   & 16.752    & 0.780 \\
d    &\checkmark&\checkmark&\checkmark&$\times$  &\checkmark   & 17.548    & 0.794 \\
e    &\checkmark&\checkmark&\checkmark&\checkmark&$\times$     & 17.564    & 0.796 \\
f    &\checkmark&\checkmark&\checkmark&\checkmark&\checkmark   & 17.602    & 0.802 \\
\hline
\end{tabular}
\vspace{-0.8em}
\end{table}

\subsection{Ablation Studies}
In this section, we perform ablation study experiments to evaluate the effectiveness of major components of our method. Specifically, we denote our frequency structure loss, frequency style loss, physical DCP loss, and physical CAP loss as "Str", "Sty", "DCP", and "CAP", respectively. Different models are denoted as follows: (a) without DRN module; (b) without structure loss; (c) without style loss; (d) without DCP loss; (e) without CAP loss; (f) our method. 
Table \ref{tab:ablation} summarizes the PSNR and SSIM results of our method under the above settings. It is obvious that our designed DRN module, Style loss, and Structure loss are of greater importance. We further verify the rationality of the DRN module in the supplementary material.

\section{Conclusion}
In this paper, we present a novel Source-Free Unsupervised Domain Adaptation image dehazing paradigm, in which only a well-trained source model and an unlabeled target
real hazy dataset are available. Our SFUDA achieves domain adaptation from the domain representation perspective and can be directly applied to existing dehazing models in a plug-and-play fashion. Specifically, we devise a Domain Representation Normalization(DRN) module to make the representation of real hazy data match the synthetic hazy domain. Besides, we leverage the frequency property and physical prior knowledge to form unsupervised losses. Extensive experiments validate that the proposed method achieves SOTA domain adaptation performances in a source-free and easy-to-use way.

\begin{acks}
This work was supported by the JKW Research Funds under Grant 20-163-14-LZ-001-004-01. We acknowledge the support of GPU cluster built by MCC Lab of Information Science and Technology Institution, USTC.
\end{acks}

\clearpage
\bibliographystyle{ACM-Reference-Format}
\bibliography{reference}


\begin{thebibliography}{61}


\ifx \showCODEN    \undefined \def \showCODEN     #1{\unskip}     \fi
\ifx \showDOI      \undefined \def \showDOI       #1{#1}\fi
\ifx \showISBNx    \undefined \def \showISBNx     #1{\unskip}     \fi
\ifx \showISBNxiii \undefined \def \showISBNxiii  #1{\unskip}     \fi
\ifx \showISSN     \undefined \def \showISSN      #1{\unskip}     \fi
\ifx \showLCCN     \undefined \def \showLCCN      #1{\unskip}     \fi
\ifx \shownote     \undefined \def \shownote      #1{#1}          \fi
\ifx \showarticletitle \undefined \def \showarticletitle #1{#1}   \fi
\ifx \showURL      \undefined \def \showURL       {\relax}        \fi
\providecommand\bibfield[2]{#2}
\providecommand\bibinfo[2]{#2}
\providecommand\natexlab[1]{#1}
\providecommand\showeprint[2][]{arXiv:#2}

\bibitem[\protect\citeauthoryear{Ahmed, Raychaudhuri, Paul, Oymak, and
  Roy-Chowdhury}{Ahmed et~al\mbox{.}}{2021}]%
        {ahmed2021unsupervised}
\bibfield{author}{\bibinfo{person}{Sk~Miraj Ahmed}, \bibinfo{person}{Dripta~S
  Raychaudhuri}, \bibinfo{person}{Sujoy Paul}, \bibinfo{person}{Samet Oymak},
  {and} \bibinfo{person}{Amit~K Roy-Chowdhury}.}
  \bibinfo{year}{2021}\natexlab{}.
\newblock \showarticletitle{Unsupervised multi-source domain adaptation without
  access to source data}. In \bibinfo{booktitle}{\emph{Proceedings of the
  IEEE/CVF Conference on Computer Vision and Pattern Recognition}}.
  \bibinfo{pages}{10103--10112}.
\newblock


\bibitem[\protect\citeauthoryear{Ancuti, Ancuti, Timofte, and
  Vleeschouwer}{Ancuti et~al\mbox{.}}{2018}]%
        {ancuti2018haze}
\bibfield{author}{\bibinfo{person}{Cosmin Ancuti}, \bibinfo{person}{Codruta~O
  Ancuti}, \bibinfo{person}{Radu Timofte}, {and} \bibinfo{person}{Christophe~De
  Vleeschouwer}.} \bibinfo{year}{2018}\natexlab{}.
\newblock \showarticletitle{I-HAZE: a dehazing benchmark with real hazy and
  haze-free indoor images}. In \bibinfo{booktitle}{\emph{International
  Conference on Advanced Concepts for Intelligent Vision Systems}}. Springer,
  \bibinfo{pages}{620--631}.
\newblock


\bibitem[\protect\citeauthoryear{Berman, Avidan, et~al\mbox{.}}{Berman
  et~al\mbox{.}}{2016}]%
        {berman2016non}
\bibfield{author}{\bibinfo{person}{Dana Berman}, \bibinfo{person}{Shai Avidan},
  {et~al\mbox{.}}} \bibinfo{year}{2016}\natexlab{}.
\newblock \showarticletitle{Non-local image dehazing}. In
  \bibinfo{booktitle}{\emph{Proceedings of the IEEE/CVF Conference on Computer
  Vision and Pattern Recognition}}. \bibinfo{pages}{1674--1682}.
\newblock


\bibitem[\protect\citeauthoryear{Cai, Xu, Jia, Qing, and Tao}{Cai
  et~al\mbox{.}}{2016}]%
        {cai2016dehazenet}
\bibfield{author}{\bibinfo{person}{Bolun Cai}, \bibinfo{person}{Xiangmin Xu},
  \bibinfo{person}{Kui Jia}, \bibinfo{person}{Chunmei Qing}, {and}
  \bibinfo{person}{Dacheng Tao}.} \bibinfo{year}{2016}\natexlab{}.
\newblock \showarticletitle{Dehazenet: An end-to-end system for single image
  haze removal}.
\newblock \bibinfo{journal}{\emph{IEEE Transactions on Image Processing}}
  \bibinfo{volume}{25}, \bibinfo{number}{11} (\bibinfo{year}{2016}),
  \bibinfo{pages}{5187--5198}.
\newblock


\bibitem[\protect\citeauthoryear{Chen, Do, and Wang}{Chen
  et~al\mbox{.}}{2016}]%
        {chen2016robust}
\bibfield{author}{\bibinfo{person}{Chen Chen}, \bibinfo{person}{Minh~N Do},
  {and} \bibinfo{person}{Jue Wang}.} \bibinfo{year}{2016}\natexlab{}.
\newblock \showarticletitle{Robust image and video dehazing with visual
  artifact suppression via gradient residual minimization}. In
  \bibinfo{booktitle}{\emph{Proceedings of the European Conference on Computer
  Vision (ECCV)}}. Springer, \bibinfo{pages}{576--591}.
\newblock


\bibitem[\protect\citeauthoryear{Chen, Wang, Yang, and Liu}{Chen
  et~al\mbox{.}}{2021}]%
        {chen2021psd}
\bibfield{author}{\bibinfo{person}{Zeyuan Chen}, \bibinfo{person}{Yangchao
  Wang}, \bibinfo{person}{Yang Yang}, {and} \bibinfo{person}{Dong Liu}.}
  \bibinfo{year}{2021}\natexlab{}.
\newblock \showarticletitle{PSD: Principled synthetic-to-real dehazing guided
  by physical priors}. In \bibinfo{booktitle}{\emph{Proceedings of the IEEE/CVF
  Conference on Computer Vision and Pattern Recognition}}.
  \bibinfo{pages}{7180--7189}.
\newblock


\bibitem[\protect\citeauthoryear{Chidlovskii, Clinchant, and
  Csurka}{Chidlovskii et~al\mbox{.}}{2016}]%
        {chidlovskii2016domain}
\bibfield{author}{\bibinfo{person}{Boris Chidlovskii},
  \bibinfo{person}{Stephane Clinchant}, {and} \bibinfo{person}{Gabriela
  Csurka}.} \bibinfo{year}{2016}\natexlab{}.
\newblock \showarticletitle{Domain adaptation in the absence of source domain
  data}. In \bibinfo{booktitle}{\emph{Proceedings of the 22nd ACM SIGKDD
  International Conference on Knowledge Discovery and Data Mining}}.
  \bibinfo{pages}{451--460}.
\newblock


\bibitem[\protect\citeauthoryear{Deng, Huang, Tsai, and Lin}{Deng
  et~al\mbox{.}}{2020}]%
        {deng2020hardgan}
\bibfield{author}{\bibinfo{person}{Qili Deng}, \bibinfo{person}{Ziling Huang},
  \bibinfo{person}{Chung-Chi Tsai}, {and} \bibinfo{person}{Chia-Wen Lin}.}
  \bibinfo{year}{2020}\natexlab{}.
\newblock \showarticletitle{Hardgan: A haze-aware representation distillation
  gan for single image dehazing}. In \bibinfo{booktitle}{\emph{Proceedings of
  the European Conference on Computer Vision (ECCV)}}. Springer,
  \bibinfo{pages}{722--738}.
\newblock


\bibitem[\protect\citeauthoryear{Dong, Pan, Xiang, Hu, Zhang, Wang, and
  Yang}{Dong et~al\mbox{.}}{2020}]%
        {dong2020multi}
\bibfield{author}{\bibinfo{person}{Hang Dong}, \bibinfo{person}{Jinshan Pan},
  \bibinfo{person}{Lei Xiang}, \bibinfo{person}{Zhe Hu}, \bibinfo{person}{Xinyi
  Zhang}, \bibinfo{person}{Fei Wang}, {and} \bibinfo{person}{Ming-Hsuan Yang}.}
  \bibinfo{year}{2020}\natexlab{}.
\newblock \showarticletitle{Multi-scale boosted dehazing network with dense
  feature fusion}. In \bibinfo{booktitle}{\emph{Proceedings of the IEEE/CVF
  Conference on Computer Vision and Pattern Recognition}}.
  \bibinfo{pages}{2157--2167}.
\newblock


\bibitem[\protect\citeauthoryear{Dong and Pan}{Dong and Pan}{2020}]%
        {dong2020physics}
\bibfield{author}{\bibinfo{person}{Jiangxin Dong} {and}
  \bibinfo{person}{Jinshan Pan}.} \bibinfo{year}{2020}\natexlab{}.
\newblock \showarticletitle{Physics-based feature dehazing networks}. In
  \bibinfo{booktitle}{\emph{Proceedings of the European Conference on Computer
  Vision (ECCV)}}. Springer, \bibinfo{pages}{188--204}.
\newblock


\bibitem[\protect\citeauthoryear{Fan, Wang, Ke, Yang, Gong, and Zhou}{Fan
  et~al\mbox{.}}{2021}]%
        {fan2021adversarially}
\bibfield{author}{\bibinfo{person}{Xinjie Fan}, \bibinfo{person}{Qifei Wang},
  \bibinfo{person}{Junjie Ke}, \bibinfo{person}{Feng Yang},
  \bibinfo{person}{Boqing Gong}, {and} \bibinfo{person}{Mingyuan Zhou}.}
  \bibinfo{year}{2021}\natexlab{}.
\newblock \showarticletitle{Adversarially adaptive normalization for single
  domain generalization}. In \bibinfo{booktitle}{\emph{Proceedings of the
  IEEE/CVF Conference on Computer Vision and Pattern Recognition}}.
  \bibinfo{pages}{8208--8217}.
\newblock


\bibitem[\protect\citeauthoryear{Fattal}{Fattal}{2008}]%
        {fattal2008single}
\bibfield{author}{\bibinfo{person}{Raanan Fattal}.}
  \bibinfo{year}{2008}\natexlab{}.
\newblock \showarticletitle{Single image dehazing}.
\newblock \bibinfo{journal}{\emph{ACM Transactions on Graphics (TOG)}}
  \bibinfo{volume}{27}, \bibinfo{number}{3} (\bibinfo{year}{2008}),
  \bibinfo{pages}{1--9}.
\newblock


\bibitem[\protect\citeauthoryear{Fattal}{Fattal}{2014}]%
        {fattal2014dehazing}
\bibfield{author}{\bibinfo{person}{Raanan Fattal}.}
  \bibinfo{year}{2014}\natexlab{}.
\newblock \showarticletitle{Dehazing using color-lines}.
\newblock \bibinfo{journal}{\emph{ACM Transactions on Graphics (TOG)}}
  \bibinfo{volume}{34}, \bibinfo{number}{1} (\bibinfo{year}{2014}),
  \bibinfo{pages}{1--14}.
\newblock


\bibitem[\protect\citeauthoryear{He, Sun, and Tang}{He et~al\mbox{.}}{2010}]%
        {he2010single}
\bibfield{author}{\bibinfo{person}{Kaiming He}, \bibinfo{person}{Jian Sun},
  {and} \bibinfo{person}{Xiaoou Tang}.} \bibinfo{year}{2010}\natexlab{}.
\newblock \showarticletitle{Single image haze removal using dark channel
  prior}.
\newblock \bibinfo{journal}{\emph{IEEE Transactions on Pattern Analysis and
  Machine Intelligence}} \bibinfo{volume}{33}, \bibinfo{number}{12}
  (\bibinfo{year}{2010}), \bibinfo{pages}{2341--2353}.
\newblock


\bibitem[\protect\citeauthoryear{He, Zhang, Zhang, Zhang, Xie, and Li}{He
  et~al\mbox{.}}{2019}]%
        {he2019bag}
\bibfield{author}{\bibinfo{person}{Tong He}, \bibinfo{person}{Zhi Zhang},
  \bibinfo{person}{Hang Zhang}, \bibinfo{person}{Zhongyue Zhang},
  \bibinfo{person}{Junyuan Xie}, {and} \bibinfo{person}{Mu Li}.}
  \bibinfo{year}{2019}\natexlab{}.
\newblock \showarticletitle{Bag of tricks for image classification with
  convolutional neural networks}. In \bibinfo{booktitle}{\emph{Proceedings of
  the IEEE/CVF Conference on Computer Vision and Pattern Recognition}}.
  \bibinfo{pages}{558--567}.
\newblock


\bibitem[\protect\citeauthoryear{Hong, Xie, Li, and Qu}{Hong
  et~al\mbox{.}}{2020}]%
        {hong2020distilling}
\bibfield{author}{\bibinfo{person}{Ming Hong}, \bibinfo{person}{Yuan Xie},
  \bibinfo{person}{Cuihua Li}, {and} \bibinfo{person}{Yanyun Qu}.}
  \bibinfo{year}{2020}\natexlab{}.
\newblock \showarticletitle{Distilling image dehazing with heterogeneous task
  imitation}. In \bibinfo{booktitle}{\emph{Proceedings of the IEEE/CVF
  Conference on Computer Vision and Pattern Recognition}}.
  \bibinfo{pages}{3462--3471}.
\newblock


\bibitem[\protect\citeauthoryear{Huang, Guan, Xiao, and Lu}{Huang
  et~al\mbox{.}}{2021}]%
        {huang2021model}
\bibfield{author}{\bibinfo{person}{Jiaxing Huang}, \bibinfo{person}{Dayan
  Guan}, \bibinfo{person}{Aoran Xiao}, {and} \bibinfo{person}{Shijian Lu}.}
  \bibinfo{year}{2021}\natexlab{}.
\newblock \showarticletitle{Model adaptation: Historical contrastive learning
  for unsupervised domain adaptation without source data}.
\newblock \bibinfo{journal}{\emph{Advances in Neural Information Processing
  Systems}}  \bibinfo{volume}{34} (\bibinfo{year}{2021}).
\newblock


\bibitem[\protect\citeauthoryear{Huang and Belongie}{Huang and
  Belongie}{2017}]%
        {huang2017arbitrary}
\bibfield{author}{\bibinfo{person}{Xun Huang} {and} \bibinfo{person}{Serge
  Belongie}.} \bibinfo{year}{2017}\natexlab{}.
\newblock \showarticletitle{Arbitrary style transfer in real-time with adaptive
  instance normalization}. In \bibinfo{booktitle}{\emph{Proceedings of the IEEE
  International Conference on Computer Vision}}. \bibinfo{pages}{1501--1510}.
\newblock


\bibitem[\protect\citeauthoryear{Jia, Chen, and Chen}{Jia
  et~al\mbox{.}}{2019}]%
        {jia2019instance}
\bibfield{author}{\bibinfo{person}{Songhao Jia}, \bibinfo{person}{Ding-Jie
  Chen}, {and} \bibinfo{person}{Hwann-Tzong Chen}.}
  \bibinfo{year}{2019}\natexlab{}.
\newblock \showarticletitle{Instance-level meta normalization}. In
  \bibinfo{booktitle}{\emph{Proceedings of the IEEE/CVF Conference on Computer
  Vision and Pattern Recognition}}. \bibinfo{pages}{4865--4873}.
\newblock


\bibitem[\protect\citeauthoryear{Jin, Lan, Zeng, Chen, and Zhang}{Jin
  et~al\mbox{.}}{2020}]%
        {jin2020style}
\bibfield{author}{\bibinfo{person}{Xin Jin}, \bibinfo{person}{Cuiling Lan},
  \bibinfo{person}{Wenjun Zeng}, \bibinfo{person}{Zhibo Chen}, {and}
  \bibinfo{person}{Li Zhang}.} \bibinfo{year}{2020}\natexlab{}.
\newblock \showarticletitle{Style normalization and restitution for
  generalizable person re-identification}. In
  \bibinfo{booktitle}{\emph{Proceedings of the IEEE/CVF Conference on Computer
  Vision and Pattern Recognition}}. \bibinfo{pages}{3143--3152}.
\newblock


\bibitem[\protect\citeauthoryear{Li, Gou, Gu, Liu, Zhou, and Peng}{Li
  et~al\mbox{.}}{2021}]%
        {li2021you}
\bibfield{author}{\bibinfo{person}{Boyun Li}, \bibinfo{person}{Yuanbiao Gou},
  \bibinfo{person}{Shuhang Gu}, \bibinfo{person}{Jerry~Zitao Liu},
  \bibinfo{person}{Joey~Tianyi Zhou}, {and} \bibinfo{person}{Xi Peng}.}
  \bibinfo{year}{2021}\natexlab{}.
\newblock \showarticletitle{You only look yourself: Unsupervised and untrained
  single image dehazing neural network}.
\newblock \bibinfo{journal}{\emph{International Journal of Computer Vision}}
  \bibinfo{volume}{129}, \bibinfo{number}{5} (\bibinfo{year}{2021}),
  \bibinfo{pages}{1754--1767}.
\newblock


\bibitem[\protect\citeauthoryear{Li, Gou, Liu, Zhu, Zhou, and Peng}{Li
  et~al\mbox{.}}{2020a}]%
        {li2020zero}
\bibfield{author}{\bibinfo{person}{Boyun Li}, \bibinfo{person}{Yuanbiao Gou},
  \bibinfo{person}{Jerry~Zitao Liu}, \bibinfo{person}{Hongyuan Zhu},
  \bibinfo{person}{Joey~Tianyi Zhou}, {and} \bibinfo{person}{Xi Peng}.}
  \bibinfo{year}{2020}\natexlab{a}.
\newblock \showarticletitle{Zero-shot image dehazing}.
\newblock \bibinfo{journal}{\emph{IEEE Transactions on Image Processing}}
  \bibinfo{volume}{29} (\bibinfo{year}{2020}), \bibinfo{pages}{8457--8466}.
\newblock


\bibitem[\protect\citeauthoryear{Li, Peng, Wang, Xu, and Feng}{Li
  et~al\mbox{.}}{2017}]%
        {li2017aod}
\bibfield{author}{\bibinfo{person}{Boyi Li}, \bibinfo{person}{Xiulian Peng},
  \bibinfo{person}{Zhangyang Wang}, \bibinfo{person}{Jizheng Xu}, {and}
  \bibinfo{person}{Dan Feng}.} \bibinfo{year}{2017}\natexlab{}.
\newblock \showarticletitle{Aod-net: All-in-one dehazing network}. In
  \bibinfo{booktitle}{\emph{Proceedings of the IEEE International Conference on
  Computer Vision}}. \bibinfo{pages}{4770--4778}.
\newblock


\bibitem[\protect\citeauthoryear{Li, Ren, Fu, Tao, Feng, Zeng, and Wang}{Li
  et~al\mbox{.}}{2018b}]%
        {li2018benchmarking}
\bibfield{author}{\bibinfo{person}{Boyi Li}, \bibinfo{person}{Wenqi Ren},
  \bibinfo{person}{Dengpan Fu}, \bibinfo{person}{Dacheng Tao},
  \bibinfo{person}{Dan Feng}, \bibinfo{person}{Wenjun Zeng}, {and}
  \bibinfo{person}{Zhangyang Wang}.} \bibinfo{year}{2018}\natexlab{b}.
\newblock \showarticletitle{Benchmarking single-image dehazing and beyond}.
\newblock \bibinfo{journal}{\emph{IEEE Transactions on Image Processing}}
  \bibinfo{volume}{28}, \bibinfo{number}{1} (\bibinfo{year}{2018}),
  \bibinfo{pages}{492--505}.
\newblock


\bibitem[\protect\citeauthoryear{Li, Dong, Ren, Pan, Gao, Sang, and Yang}{Li
  et~al\mbox{.}}{2019b}]%
        {li2019semi}
\bibfield{author}{\bibinfo{person}{Lerenhan Li}, \bibinfo{person}{Yunlong
  Dong}, \bibinfo{person}{Wenqi Ren}, \bibinfo{person}{Jinshan Pan},
  \bibinfo{person}{Changxin Gao}, \bibinfo{person}{Nong Sang}, {and}
  \bibinfo{person}{Ming-Hsuan Yang}.} \bibinfo{year}{2019}\natexlab{b}.
\newblock \showarticletitle{Semi-supervised image dehazing}.
\newblock \bibinfo{journal}{\emph{IEEE Transactions on Image Processing}}
  \bibinfo{volume}{29} (\bibinfo{year}{2019}), \bibinfo{pages}{2766--2779}.
\newblock


\bibitem[\protect\citeauthoryear{Li, Jiao, Cao, Wong, and Wu}{Li
  et~al\mbox{.}}{2020b}]%
        {li2020model}
\bibfield{author}{\bibinfo{person}{Rui Li}, \bibinfo{person}{Qianfen Jiao},
  \bibinfo{person}{Wenming Cao}, \bibinfo{person}{Hau-San Wong}, {and}
  \bibinfo{person}{Si Wu}.} \bibinfo{year}{2020}\natexlab{b}.
\newblock \showarticletitle{Model adaptation: Unsupervised domain adaptation
  without source data}. In \bibinfo{booktitle}{\emph{Proceedings of the
  IEEE/CVF Conference on Computer Vision and Pattern Recognition}}.
  \bibinfo{pages}{9641--9650}.
\newblock


\bibitem[\protect\citeauthoryear{Li, Pan, Li, and Tang}{Li
  et~al\mbox{.}}{2018a}]%
        {li2018single}
\bibfield{author}{\bibinfo{person}{Runde Li}, \bibinfo{person}{Jinshan Pan},
  \bibinfo{person}{Zechao Li}, {and} \bibinfo{person}{Jinhui Tang}.}
  \bibinfo{year}{2018}\natexlab{a}.
\newblock \showarticletitle{Single image dehazing via conditional generative
  adversarial network}. In \bibinfo{booktitle}{\emph{Proceedings of the IEEE
  Conference on Computer Vision and Pattern Recognition}}.
  \bibinfo{pages}{8202--8211}.
\newblock


\bibitem[\protect\citeauthoryear{Li, Araujo, Ren, Wang, Tokuda, Junior,
  Cesar-Junior, Zhang, Guo, and Cao}{Li et~al\mbox{.}}{2019a}]%
        {li2019single}
\bibfield{author}{\bibinfo{person}{Siyuan Li}, \bibinfo{person}{Iago~Breno
  Araujo}, \bibinfo{person}{Wenqi Ren}, \bibinfo{person}{Zhangyang Wang},
  \bibinfo{person}{Eric~K Tokuda}, \bibinfo{person}{Roberto~Hirata Junior},
  \bibinfo{person}{Roberto Cesar-Junior}, \bibinfo{person}{Jiawan Zhang},
  \bibinfo{person}{Xiaojie Guo}, {and} \bibinfo{person}{Xiaochun Cao}.}
  \bibinfo{year}{2019}\natexlab{a}.
\newblock \showarticletitle{Single image deraining: A comprehensive benchmark
  analysis}. In \bibinfo{booktitle}{\emph{Proceedings of the IEEE/CVF
  Conference on Computer Vision and Pattern Recognition}}.
  \bibinfo{pages}{3838--3847}.
\newblock


\bibitem[\protect\citeauthoryear{Liu, Fan, Hou, Jiang, Luo, and Zhang}{Liu
  et~al\mbox{.}}{2018}]%
        {liu2018learning}
\bibfield{author}{\bibinfo{person}{Risheng Liu}, \bibinfo{person}{Xin Fan},
  \bibinfo{person}{Minjun Hou}, \bibinfo{person}{Zhiying Jiang},
  \bibinfo{person}{Zhongxuan Luo}, {and} \bibinfo{person}{Lei Zhang}.}
  \bibinfo{year}{2018}\natexlab{}.
\newblock \showarticletitle{Learning aggregated transmission propagation
  networks for haze removal and beyond}.
\newblock \bibinfo{journal}{\emph{IEEE Transactions on Neural Networks and
  Learning Systems}} \bibinfo{volume}{30}, \bibinfo{number}{10}
  (\bibinfo{year}{2018}), \bibinfo{pages}{2973--2986}.
\newblock


\bibitem[\protect\citeauthoryear{Liu, Ma, Shi, and Chen}{Liu
  et~al\mbox{.}}{2019a}]%
        {liu2019griddehazenet}
\bibfield{author}{\bibinfo{person}{Xiaohong Liu}, \bibinfo{person}{Yongrui Ma},
  \bibinfo{person}{Zhihao Shi}, {and} \bibinfo{person}{Jun Chen}.}
  \bibinfo{year}{2019}\natexlab{a}.
\newblock \showarticletitle{Griddehazenet: Attention-based multi-scale network
  for image dehazing}. In \bibinfo{booktitle}{\emph{Proceedings of the IEEE/CVF
  International Conference on Computer Vision}}. \bibinfo{pages}{7314--7323}.
\newblock


\bibitem[\protect\citeauthoryear{Liu, Pan, Ren, and Su}{Liu
  et~al\mbox{.}}{2019b}]%
        {liu2019learning}
\bibfield{author}{\bibinfo{person}{Yang Liu}, \bibinfo{person}{Jinshan Pan},
  \bibinfo{person}{Jimmy Ren}, {and} \bibinfo{person}{Zhixun Su}.}
  \bibinfo{year}{2019}\natexlab{b}.
\newblock \showarticletitle{Learning deep priors for image dehazing}. In
  \bibinfo{booktitle}{\emph{Proceedings of the IEEE/CVF International
  Conference on Computer Vision}}. \bibinfo{pages}{2492--2500}.
\newblock


\bibitem[\protect\citeauthoryear{Liu, Zhang, and Wang}{Liu
  et~al\mbox{.}}{2021a}]%
        {liu2021source}
\bibfield{author}{\bibinfo{person}{Yuang Liu}, \bibinfo{person}{Wei Zhang},
  {and} \bibinfo{person}{Jun Wang}.} \bibinfo{year}{2021}\natexlab{a}.
\newblock \showarticletitle{Source-free domain adaptation for semantic
  segmentation}. In \bibinfo{booktitle}{\emph{Proceedings of the IEEE/CVF
  Conference on Computer Vision and Pattern Recognition}}.
  \bibinfo{pages}{1215--1224}.
\newblock


\bibitem[\protect\citeauthoryear{Liu, Zhu, Pei, Fu, Qin, Zhang, Wan, and
  Feng}{Liu et~al\mbox{.}}{2021b}]%
        {liu2021synthetic}
\bibfield{author}{\bibinfo{person}{Ye Liu}, \bibinfo{person}{Lei Zhu},
  \bibinfo{person}{Shunda Pei}, \bibinfo{person}{Huazhu Fu},
  \bibinfo{person}{Jing Qin}, \bibinfo{person}{Qing Zhang},
  \bibinfo{person}{Liang Wan}, {and} \bibinfo{person}{Wei Feng}.}
  \bibinfo{year}{2021}\natexlab{b}.
\newblock \showarticletitle{From Synthetic to Real: Image Dehazing
  Collaborating with Unlabeled Real Data}. In
  \bibinfo{booktitle}{\emph{Proceedings of the 29th ACM International
  Conference on Multimedia}}. \bibinfo{pages}{50--58}.
\newblock


\bibitem[\protect\citeauthoryear{McCartney}{McCartney}{1976}]%
        {mccartney1976optics}
\bibfield{author}{\bibinfo{person}{Earl~J McCartney}.}
  \bibinfo{year}{1976}\natexlab{}.
\newblock \showarticletitle{Optics of the atmosphere: scattering by molecules
  and particles}.
\newblock \bibinfo{journal}{\emph{New York}} (\bibinfo{year}{1976}).
\newblock


\bibitem[\protect\citeauthoryear{Mittal, Moorthy, and Bovik}{Mittal
  et~al\mbox{.}}{2012a}]%
        {mittal2012no}
\bibfield{author}{\bibinfo{person}{Anish Mittal},
  \bibinfo{person}{Anush~Krishna Moorthy}, {and} \bibinfo{person}{Alan~Conrad
  Bovik}.} \bibinfo{year}{2012}\natexlab{a}.
\newblock \showarticletitle{No-reference image quality assessment in the
  spatial domain}.
\newblock \bibinfo{journal}{\emph{IEEE Transactions on Image Processing}}
  \bibinfo{volume}{21}, \bibinfo{number}{12} (\bibinfo{year}{2012}),
  \bibinfo{pages}{4695--4708}.
\newblock


\bibitem[\protect\citeauthoryear{Mittal, Soundararajan, and Bovik}{Mittal
  et~al\mbox{.}}{2012b}]%
        {mittal2012making}
\bibfield{author}{\bibinfo{person}{Anish Mittal}, \bibinfo{person}{Rajiv
  Soundararajan}, {and} \bibinfo{person}{Alan~C Bovik}.}
  \bibinfo{year}{2012}\natexlab{b}.
\newblock \showarticletitle{Making a “completely blind” image quality
  analyzer}.
\newblock \bibinfo{journal}{\emph{IEEE Signal Processing Letters}}
  \bibinfo{volume}{20}, \bibinfo{number}{3} (\bibinfo{year}{2012}),
  \bibinfo{pages}{209--212}.
\newblock


\bibitem[\protect\citeauthoryear{Oppenheim and Lim}{Oppenheim and Lim}{1981}]%
        {oppenheim1981importance}
\bibfield{author}{\bibinfo{person}{Alan~V Oppenheim} {and}
  \bibinfo{person}{Jae~S Lim}.} \bibinfo{year}{1981}\natexlab{}.
\newblock \showarticletitle{The importance of phase in signals}.
\newblock \bibinfo{journal}{\emph{Proc. IEEE}} \bibinfo{volume}{69},
  \bibinfo{number}{5} (\bibinfo{year}{1981}), \bibinfo{pages}{529--541}.
\newblock


\bibitem[\protect\citeauthoryear{Pan, Luo, Shi, and Tang}{Pan
  et~al\mbox{.}}{2018}]%
        {pan2018two}
\bibfield{author}{\bibinfo{person}{Xingang Pan}, \bibinfo{person}{Ping Luo},
  \bibinfo{person}{Jianping Shi}, {and} \bibinfo{person}{Xiaoou Tang}.}
  \bibinfo{year}{2018}\natexlab{}.
\newblock \showarticletitle{Two at once: Enhancing learning and generalization
  capacities via ibn-net}. In \bibinfo{booktitle}{\emph{Proceedings of the
  European Conference on Computer Vision (ECCV)}}. \bibinfo{pages}{464--479}.
\newblock


\bibitem[\protect\citeauthoryear{Pan, Li, He, Yao, Wu, Lin, Li, and Ding}{Pan
  et~al\mbox{.}}{2022}]%
        {pan2022towards}
\bibfield{author}{\bibinfo{person}{Zhihong Pan}, \bibinfo{person}{Baopu Li},
  \bibinfo{person}{Dongliang He}, \bibinfo{person}{Mingde Yao},
  \bibinfo{person}{Wenhao Wu}, \bibinfo{person}{Tianwei Lin},
  \bibinfo{person}{Xin Li}, {and} \bibinfo{person}{Errui Ding}.}
  \bibinfo{year}{2022}\natexlab{}.
\newblock \showarticletitle{Towards Bidirectional Arbitrary Image Rescaling:
  Joint Optimization and Cycle Idempotence}. In
  \bibinfo{booktitle}{\emph{Proceedings of the IEEE/CVF Conference on Computer
  Vision and Pattern Recognition}}. \bibinfo{pages}{17389--17398}.
\newblock


\bibitem[\protect\citeauthoryear{Pizer, Amburn, Austin, Cromartie, Geselowitz,
  Greer, ter Haar~Romeny, Zimmerman, and Zuiderveld}{Pizer
  et~al\mbox{.}}{1987}]%
        {pizer1987adaptive}
\bibfield{author}{\bibinfo{person}{Stephen~M Pizer}, \bibinfo{person}{E~Philip
  Amburn}, \bibinfo{person}{John~D Austin}, \bibinfo{person}{Robert Cromartie},
  \bibinfo{person}{Ari Geselowitz}, \bibinfo{person}{Trey Greer},
  \bibinfo{person}{Bart ter Haar~Romeny}, \bibinfo{person}{John~B Zimmerman},
  {and} \bibinfo{person}{Karel Zuiderveld}.} \bibinfo{year}{1987}\natexlab{}.
\newblock \showarticletitle{Adaptive histogram equalization and its
  variations}.
\newblock \bibinfo{journal}{\emph{Computer Vision, Graphics, and Image
  Processing}} \bibinfo{volume}{39}, \bibinfo{number}{3}
  (\bibinfo{year}{1987}), \bibinfo{pages}{355--368}.
\newblock


\bibitem[\protect\citeauthoryear{Qin, Wang, Bai, Xie, and Jia}{Qin
  et~al\mbox{.}}{2020}]%
        {qin2020ffa}
\bibfield{author}{\bibinfo{person}{Xu Qin}, \bibinfo{person}{Zhilin Wang},
  \bibinfo{person}{Yuanchao Bai}, \bibinfo{person}{Xiaodong Xie}, {and}
  \bibinfo{person}{Huizhu Jia}.} \bibinfo{year}{2020}\natexlab{}.
\newblock \showarticletitle{FFA-Net: Feature fusion attention network for
  single image dehazing}. In \bibinfo{booktitle}{\emph{Proceedings of the AAAI
  Conference on Artificial Intelligence}}, Vol.~\bibinfo{volume}{34}.
  \bibinfo{pages}{11908--11915}.
\newblock


\bibitem[\protect\citeauthoryear{Qu, Chen, Huang, and Xie}{Qu
  et~al\mbox{.}}{2019}]%
        {qu2019enhanced}
\bibfield{author}{\bibinfo{person}{Yanyun Qu}, \bibinfo{person}{Yizi Chen},
  \bibinfo{person}{Jingying Huang}, {and} \bibinfo{person}{Yuan Xie}.}
  \bibinfo{year}{2019}\natexlab{}.
\newblock \showarticletitle{Enhanced pix2pix dehazing network}. In
  \bibinfo{booktitle}{\emph{Proceedings of the IEEE/CVF Conference on Computer
  Vision and Pattern Recognition}}. \bibinfo{pages}{8160--8168}.
\newblock


\bibitem[\protect\citeauthoryear{Ren, Liu, Zhang, Pan, Cao, and Yang}{Ren
  et~al\mbox{.}}{2016}]%
        {ren2016single}
\bibfield{author}{\bibinfo{person}{Wenqi Ren}, \bibinfo{person}{Si Liu},
  \bibinfo{person}{Hua Zhang}, \bibinfo{person}{Jinshan Pan},
  \bibinfo{person}{Xiaochun Cao}, {and} \bibinfo{person}{Ming-Hsuan Yang}.}
  \bibinfo{year}{2016}\natexlab{}.
\newblock \showarticletitle{Single image dehazing via multi-scale convolutional
  neural networks}. In \bibinfo{booktitle}{\emph{Proceedings of the European
  Conference on Computer Vision (ECCV)}}. Springer, \bibinfo{pages}{154--169}.
\newblock


\bibitem[\protect\citeauthoryear{Ren, Ma, Zhang, Pan, Cao, Liu, and Yang}{Ren
  et~al\mbox{.}}{2018}]%
        {ren2018gated}
\bibfield{author}{\bibinfo{person}{Wenqi Ren}, \bibinfo{person}{Lin Ma},
  \bibinfo{person}{Jiawei Zhang}, \bibinfo{person}{Jinshan Pan},
  \bibinfo{person}{Xiaochun Cao}, \bibinfo{person}{Wei Liu}, {and}
  \bibinfo{person}{Ming-Hsuan Yang}.} \bibinfo{year}{2018}\natexlab{}.
\newblock \showarticletitle{Gated fusion network for single image dehazing}. In
  \bibinfo{booktitle}{\emph{Proceedings of the IEEE Conference on Computer
  Vision and Pattern Recognition}}. \bibinfo{pages}{3253--3261}.
\newblock


\bibitem[\protect\citeauthoryear{Reza}{Reza}{2004}]%
        {reza2004realization}
\bibfield{author}{\bibinfo{person}{Ali~M Reza}.}
  \bibinfo{year}{2004}\natexlab{}.
\newblock \showarticletitle{Realization of the contrast limited adaptive
  histogram equalization (CLAHE) for real-time image enhancement}.
\newblock \bibinfo{journal}{\emph{Journal of Signal Processing Systems for
  Signal Image and Video Technology}} \bibinfo{volume}{38}, \bibinfo{number}{1}
  (\bibinfo{year}{2004}), \bibinfo{pages}{35--44}.
\newblock


\bibitem[\protect\citeauthoryear{Shao, Li, Ren, Gao, and Sang}{Shao
  et~al\mbox{.}}{2020}]%
        {shao2020domain}
\bibfield{author}{\bibinfo{person}{Yuanjie Shao}, \bibinfo{person}{Lerenhan
  Li}, \bibinfo{person}{Wenqi Ren}, \bibinfo{person}{Changxin Gao}, {and}
  \bibinfo{person}{Nong Sang}.} \bibinfo{year}{2020}\natexlab{}.
\newblock \showarticletitle{Domain adaptation for image dehazing}. In
  \bibinfo{booktitle}{\emph{Proceedings of the IEEE/CVF Conference on Computer
  Vision and Pattern Recognition}}. \bibinfo{pages}{2808--2817}.
\newblock


\bibitem[\protect\citeauthoryear{Shyam, Yoon, and Kim}{Shyam
  et~al\mbox{.}}{2021}]%
        {shyam2021towards}
\bibfield{author}{\bibinfo{person}{Pranjay Shyam}, \bibinfo{person}{Kuk-Jin
  Yoon}, {and} \bibinfo{person}{Kyung-Soo Kim}.}
  \bibinfo{year}{2021}\natexlab{}.
\newblock \showarticletitle{Towards domain invariant single image dehazing}.
\newblock \bibinfo{journal}{\emph{arXiv preprint arXiv:2101.10449}}
  (\bibinfo{year}{2021}).
\newblock


\bibitem[\protect\citeauthoryear{Skarbnik, Zeevi, and Sagiv}{Skarbnik
  et~al\mbox{.}}{2009}]%
        {skarbnik2009importance}
\bibfield{author}{\bibinfo{person}{Nikolay Skarbnik},
  \bibinfo{person}{Yehoshua~Y Zeevi}, {and} \bibinfo{person}{Chen Sagiv}.}
  \bibinfo{year}{2009}\natexlab{}.
\newblock \bibinfo{booktitle}{\emph{The importance of phase in image
  processing}}.
\newblock \bibinfo{publisher}{Technion-Israel Institute of Technology, Faculty
  of Electrical Engineering}.
\newblock


\bibitem[\protect\citeauthoryear{VidalMata, Banerjee, RichardWebster, Albright,
  Davalos, McCloskey, Miller, Tambo, Ghosh, Nagesh, et~al\mbox{.}}{VidalMata
  et~al\mbox{.}}{2020}]%
        {vidalmata2020bridging}
\bibfield{author}{\bibinfo{person}{Rosaura~G VidalMata}, \bibinfo{person}{Sreya
  Banerjee}, \bibinfo{person}{Brandon RichardWebster}, \bibinfo{person}{Michael
  Albright}, \bibinfo{person}{Pedro Davalos}, \bibinfo{person}{Scott
  McCloskey}, \bibinfo{person}{Ben Miller}, \bibinfo{person}{Asong Tambo},
  \bibinfo{person}{Sushobhan Ghosh}, \bibinfo{person}{Sudarshan Nagesh},
  {et~al\mbox{.}}} \bibinfo{year}{2020}\natexlab{}.
\newblock \showarticletitle{Bridging the gap between computational photography
  and visual recognition}.
\newblock \bibinfo{journal}{\emph{IEEE Transactions on Pattern Analysis and
  Machine Intelligence}} \bibinfo{volume}{43}, \bibinfo{number}{12}
  (\bibinfo{year}{2020}), \bibinfo{pages}{4272--4290}.
\newblock


\bibitem[\protect\citeauthoryear{Wang, Shen, Fan, Shao, Yang, Luo, and
  Deng}{Wang et~al\mbox{.}}{2021b}]%
        {wang2021eaa}
\bibfield{author}{\bibinfo{person}{Chao Wang}, \bibinfo{person}{Hao-Zhen Shen},
  \bibinfo{person}{Fan Fan}, \bibinfo{person}{Ming-Wen Shao},
  \bibinfo{person}{Chuan-Sheng Yang}, \bibinfo{person}{Jian-Cheng Luo}, {and}
  \bibinfo{person}{Liang-Jian Deng}.} \bibinfo{year}{2021}\natexlab{b}.
\newblock \showarticletitle{EAA-Net: A novel edge assisted attention network
  for single image dehazing}.
\newblock \bibinfo{journal}{\emph{Knowledge-Based Systems}}
  \bibinfo{volume}{228} (\bibinfo{year}{2021}), \bibinfo{pages}{107279}.
\newblock


\bibitem[\protect\citeauthoryear{Wang, Fu, Sun, and Zha}{Wang
  et~al\mbox{.}}{2021a}]%
        {wang2021jpeg}
\bibfield{author}{\bibinfo{person}{Menglu Wang}, \bibinfo{person}{Xueyang Fu},
  \bibinfo{person}{Zepei Sun}, {and} \bibinfo{person}{Zheng-Jun Zha}.}
  \bibinfo{year}{2021}\natexlab{a}.
\newblock \showarticletitle{JPEG artifacts removal via compression quality
  ranker-guided networks}. In \bibinfo{booktitle}{\emph{Proceedings of the
  Twenty-Ninth International Conference on International Joint Conferences on
  Artificial Intelligence}}. \bibinfo{pages}{566--572}.
\newblock


\bibitem[\protect\citeauthoryear{Wu, Qu, Lin, Zhou, Qiao, Zhang, Xie, and
  Ma}{Wu et~al\mbox{.}}{2021}]%
        {wu2021contrastive}
\bibfield{author}{\bibinfo{person}{Haiyan Wu}, \bibinfo{person}{Yanyun Qu},
  \bibinfo{person}{Shaohui Lin}, \bibinfo{person}{Jian Zhou},
  \bibinfo{person}{Ruizhi Qiao}, \bibinfo{person}{Zhizhong Zhang},
  \bibinfo{person}{Yuan Xie}, {and} \bibinfo{person}{Lizhuang Ma}.}
  \bibinfo{year}{2021}\natexlab{}.
\newblock \showarticletitle{Contrastive Learning for Compact Single Image
  Dehazing}. In \bibinfo{booktitle}{\emph{Proceedings of the IEEE/CVF
  Conference on Computer Vision and Pattern Recognition}}.
  \bibinfo{pages}{10551--10560}.
\newblock


\bibitem[\protect\citeauthoryear{Yang, Yuan, Ren, Liu, Scheirer, Wang, Zhang,
  Zhong, Xie, Pu, et~al\mbox{.}}{Yang et~al\mbox{.}}{2020}]%
        {yang2020advancing}
\bibfield{author}{\bibinfo{person}{Wenhan Yang}, \bibinfo{person}{Ye Yuan},
  \bibinfo{person}{Wenqi Ren}, \bibinfo{person}{Jiaying Liu},
  \bibinfo{person}{Walter~J Scheirer}, \bibinfo{person}{Zhangyang Wang},
  \bibinfo{person}{Taiheng Zhang}, \bibinfo{person}{Qiaoyong Zhong},
  \bibinfo{person}{Di Xie}, \bibinfo{person}{Shiliang Pu}, {et~al\mbox{.}}}
  \bibinfo{year}{2020}\natexlab{}.
\newblock \showarticletitle{Advancing image understanding in poor visibility
  environments: A collective benchmark study}.
\newblock \bibinfo{journal}{\emph{IEEE Transactions on Image Processing}}
  \bibinfo{volume}{29} (\bibinfo{year}{2020}), \bibinfo{pages}{5737--5752}.
\newblock


\bibitem[\protect\citeauthoryear{Yao, Xiong, Wang, Liu, and Chen}{Yao
  et~al\mbox{.}}{2019}]%
        {Yao2019spectral}
\bibfield{author}{\bibinfo{person}{Mingde Yao}, \bibinfo{person}{Zhiwei Xiong},
  \bibinfo{person}{Lizhi Wang}, \bibinfo{person}{Dong Liu}, {and}
  \bibinfo{person}{Xuejin Chen}.} \bibinfo{year}{2019}\natexlab{}.
\newblock \showarticletitle{Spectral-depth imaging with deep learning based
  reconstruction}.
\newblock \bibinfo{journal}{\emph{Optics Express}} \bibinfo{volume}{27},
  \bibinfo{number}{26} (\bibinfo{year}{2019}), \bibinfo{pages}{38312--38325}.
\newblock


\bibitem[\protect\citeauthoryear{Ye, Jiang, Zhang, Chen, Chen, Chen, and Lu}{Ye
  et~al\mbox{.}}{2021}]%
        {ye2021perceiving}
\bibfield{author}{\bibinfo{person}{Tian Ye}, \bibinfo{person}{Mingchao Jiang},
  \bibinfo{person}{Yunchen Zhang}, \bibinfo{person}{Liang Chen},
  \bibinfo{person}{Erkang Chen}, \bibinfo{person}{Pen Chen}, {and}
  \bibinfo{person}{Zhiyong Lu}.} \bibinfo{year}{2021}\natexlab{}.
\newblock \showarticletitle{Perceiving and Modeling Density is All You Need for
  Image Dehazing}.
\newblock \bibinfo{journal}{\emph{arXiv preprint arXiv:2111.09733}}
  (\bibinfo{year}{2021}).
\newblock


\bibitem[\protect\citeauthoryear{Ye, Liu, Zhang, Chen, and Chen}{Ye
  et~al\mbox{.}}{2022}]%
        {ye2022mutual}
\bibfield{author}{\bibinfo{person}{Tian Ye}, \bibinfo{person}{Yun Liu},
  \bibinfo{person}{Yunchen Zhang}, \bibinfo{person}{Sixiang Chen}, {and}
  \bibinfo{person}{Erkang Chen}.} \bibinfo{year}{2022}\natexlab{}.
\newblock \showarticletitle{Mutual Learning for Domain Adaptation:
  Self-distillation Image Dehazing Network with Sample-cycle}.
\newblock \bibinfo{journal}{\emph{arXiv preprint arXiv:2203.09430}}
  (\bibinfo{year}{2022}).
\newblock


\bibitem[\protect\citeauthoryear{Zhang and Patel}{Zhang and Patel}{2018}]%
        {zhang2018densely}
\bibfield{author}{\bibinfo{person}{He Zhang} {and} \bibinfo{person}{Vishal~M
  Patel}.} \bibinfo{year}{2018}\natexlab{}.
\newblock \showarticletitle{Densely connected pyramid dehazing network}. In
  \bibinfo{booktitle}{\emph{Proceedings of the IEEE Conference on Computer
  Vision and Pattern Recognition}}. \bibinfo{pages}{3194--3203}.
\newblock


\bibitem[\protect\citeauthoryear{Zhang, Cao, Fang, Kang, and Wen~Chen}{Zhang
  et~al\mbox{.}}{2017}]%
        {zhang2017fast}
\bibfield{author}{\bibinfo{person}{Jing Zhang}, \bibinfo{person}{Yang Cao},
  \bibinfo{person}{Shuai Fang}, \bibinfo{person}{Yu Kang}, {and}
  \bibinfo{person}{Chang Wen~Chen}.} \bibinfo{year}{2017}\natexlab{}.
\newblock \showarticletitle{Fast haze removal for nighttime image using maximum
  reflectance prior}. In \bibinfo{booktitle}{\emph{Proceedings of the IEEE
  Conference on Computer Vision and Pattern Recognition}}.
  \bibinfo{pages}{7418--7426}.
\newblock


\bibitem[\protect\citeauthoryear{Zheng, Ren, Cao, Hu, Wang, Song, and
  Jia}{Zheng et~al\mbox{.}}{2021}]%
        {zheng2021ultra}
\bibfield{author}{\bibinfo{person}{Zhuoran Zheng}, \bibinfo{person}{Wenqi Ren},
  \bibinfo{person}{Xiaochun Cao}, \bibinfo{person}{Xiaobin Hu},
  \bibinfo{person}{Tao Wang}, \bibinfo{person}{Fenglong Song}, {and}
  \bibinfo{person}{Xiuyi Jia}.} \bibinfo{year}{2021}\natexlab{}.
\newblock \showarticletitle{Ultra-high-definition image dehazing via
  multi-guided bilateral learning}. In \bibinfo{booktitle}{\emph{Proceedings of
  the IEEE Conference on Computer Vision and Pattern Recognition}}. IEEE,
  \bibinfo{pages}{16180--16189}.
\newblock


\bibitem[\protect\citeauthoryear{Zhu, Mai, and Shao}{Zhu et~al\mbox{.}}{2014}]%
        {zhu2014single}
\bibfield{author}{\bibinfo{person}{Qingsong Zhu}, \bibinfo{person}{Jiaming
  Mai}, {and} \bibinfo{person}{Ling Shao}.} \bibinfo{year}{2014}\natexlab{}.
\newblock \showarticletitle{Single Image Dehazing Using Color Attenuation
  Prior.}. In \bibinfo{booktitle}{\emph{British Machine Vision
  Conference(BMVC)}}. Citeseer.
\newblock


\bibitem[\protect\citeauthoryear{Zhu, Mai, and Shao}{Zhu et~al\mbox{.}}{2015}]%
        {zhu2015fast}
\bibfield{author}{\bibinfo{person}{Qingsong Zhu}, \bibinfo{person}{Jiaming
  Mai}, {and} \bibinfo{person}{Ling Shao}.} \bibinfo{year}{2015}\natexlab{}.
\newblock \showarticletitle{A fast single image haze removal algorithm using
  color attenuation prior}.
\newblock \bibinfo{journal}{\emph{IEEE Transactions on Image Processing}}
  \bibinfo{volume}{24}, \bibinfo{number}{11} (\bibinfo{year}{2015}),
  \bibinfo{pages}{3522--3533}.
\newblock


\end{thebibliography}

\end{document}